\documentclass[%
 aip,
 cha,
 amsmath,amssymb,
 reprint,%
]{revtex4-2}

\usepackage[utf8]{inputenc}
\usepackage{amsmath,amssymb}
\usepackage{graphicx}
\usepackage{hyperref}
\usepackage{xcolor}
\usepackage{tabularx}
\usepackage{tikz}
\usepackage[percent]{overpic}
\usetikzlibrary{arrows.meta, positioning, calc}
\usepackage{subcaption}

\newcommand{\uu}{\mathbf{u}}
\newcommand{\uuH}{\mathbf{u}_H}
\newcommand{\rr}{\mathbf{r}}

\newcommand{\ff}{\mathbf{f}}
\newcommand{\MM}{\mathbf{M}}
\newcommand{\WWi}{\mathbf{W}_\text{in}}
\newcommand{\WWo}{\mathbf{W}_\text{out}}
\newcommand{\UU}{\mathbf{U}}
\newcommand{\PP}{\mathbf{P}}
\newcommand{\R}{\mathbb{R}}
\newcommand{\N}{\mathbb{N}}

\begin{document}

\title{Next-Generation Reservoir Computing for Dynamical Inference}

\author{Rok Cestnik}
\email{rok.cestnik@math.lth.se}
\affiliation{Centre for Mathematical Science, Lund University, M\"arkesbacken 4, 223 62 Lund, Sweden}

\author{Erik A. Martens}
\affiliation{Centre for Mathematical Science, Lund University, M\"arkesbacken 4, 223 62 Lund, Sweden}

\affiliation{IMFUFA, Department of Science and Environment, Roskilde University, Universitetsvej 1, Roskilde, Denmark}

\date{\today}

\begin{abstract}
We present a simple and scalable implementation of next-generation reservoir computing (NGRC) for modeling dynamical systems from time-series data. The method uses a pseudorandom nonlinear projection of time-delay embedded inputs, allowing the feature-space dimension to be chosen independently of the observation size and offering a flexible alternative to polynomial-based NGRC projections. We demonstrate the approach on benchmark tasks, including attractor reconstruction and bifurcation diagram estimation, using partial and noisy measurements. We further show that small amounts of measurement noise during training act as an effective regularizer, improving long-term autonomous stability compared to standard regression alone. Across all tests, the models remain stable over long rollouts and generalize beyond the training data. The framework offers explicit control of system state during prediction, and these properties make NGRC a natural candidate for applications such as surrogate modeling and digital‐twin applications.
\end{abstract}

\maketitle

\begin{table*}[htp!]

\begin{tabular}{l l}
\hline \hline \\
$\uu(t)\in\R^N$& System state at time $t$\\
$\uuH(t)\in\R^{H\times N}$&
Measurement vector of history-embedded system states, containing state vectors at times $t,t-\tau, \ldots t-(H-1)\cdot \tau$.
\\
$\rr\in\R^{M}$& Reservoir state at time $t$\\
$P: \uuH\mapsto \rr$& Projection of the measurement vector to the higher-dimensional reservoir space\\
$p_m\in\R$& Elements of the vector $P(\uuH)=[p_1,p_2,\ldots,p_M]$  \\
$\PP\in\R^{M\times T}$& Matrix of all projected measurement vectors for the complete dataset \\
$\WWo\in\R^{M\times N}$& Matrix of the output weights. Prediction model for the linear relation $\rr = \WWo  \uu$ \\
$\UU\in\R^{N\times T}$& Matrix of all state vectors for the complete dataset \\
$\tau\in\R^+$& Time-sampling step \\
$T\in\N$& Number of time data points \\
$N\in\N$& System state dimension $\uu(t)$ \\
$H\in\N$& Time-embedding dimension \\
$M\in\N$& Projection dimension, reservoir space dimension, $M\gg1$ \\
$E\in\R$& Mean square error of one-step predictions, quantifying success of training\\
\vspace{0.5em}\\
\hline \hline
\end{tabular}
\caption{List of symbols.}
\end{table*}

\begin{quotation}
Reservoir computing (RC)~\cite{Ott2018} is a machine learning method used to predict  the dynamic behavior of complex systems. It relies on a reservoir --- a fixed, dynamical system --- into which an input signal from a given external system of interest is injected. The output of the reservoir is then passed through a trained \emph{readout layer} in order to predict future states of the external system.
Next-generation reservoir computing (NGRC)~\cite{NGRC2021} is a simplified variant of RC that replaces the nonlinear reservoir with a direct \emph{feature expansion}.
Here, \emph{features} refer to measurable properties derived from data, and \emph{feature expansion} means generating additional features by combining existing ones.
This simplification makes NGRC more interpretable and easier to control than traditional RC.
In this work, we introduce a pseudorandom projection as a simple and scalable way to build these features, and use it to model time-series data. Our approach makes accurate and stable predictions even when observations are noisy or incomplete. Through examples such as chaotic attractor reconstruction, bifurcation mapping, and phase recovery, we illustrate how next-generation reservoir computing can capture essential system dynamics and generalize beyond the training data. Our results highlight their potential as a practical tool for applications ranging from digital twins to models of physical, biological, and financial systems.
\end{quotation}

\section{Introduction}

Reservoir computing (RC) has emerged as a powerful and versatile machine learning framework for processing temporal data, particularly in tasks involving time series prediction and the inference of latent variables in dynamical systems~\cite{Jaeger2001, maass2002, lukosevicius2009}.
As Lu {\it et al.}~\cite{Ott2018} describe, {\it ``Reservoir computing is a machine-learning approach that has demonstrated success at a variety of tasks, including time series prediction and inferring unmeasured variables of a dynamical system from measured variables''}. 
In its classical form, RC leverages a high-dimensional, nonlinear dynamical system --- referred to as the \emph{reservoir} --- which is driven by an input signal. The reservoir's response is then post-processed with a linear map to obtain the desired output, commonly referred to as \emph{readout}. For a general overview of the theory and state of the art, see also Refs.~\cite{tanaka2019recent,nakajima2021reservoir}.

A key advantage of RC lies in its simplicity: the reservoir itself operates as a static component that does not require training, while only the output layer (readout) is optimized, typically through linear regression. 
This design allows for implementations of the reservoir in both software --- often utilizing recurrent neural networks --- and physical systems. 
Such flexibility has led to a wide range of serious physical implementations, including optical systems~\cite{appeltant2011}, electronic delay circuits~\cite{haynes2015}, and photonic devices~\cite{larger2017}. And to illustrate just how general the reservoir-computing principle is, researchers have even demonstrated it using simple water tanks where surface ripples act as the reservoir~\cite{fernando2003} --- a setup meant more as a proof-of-concept than a practical device. Quantum-inspired implementations of reservoir computing have also been explored in recent work~\cite{domingo2023arxiv}.
The capacity to treat the reservoir as a black box, without needing a detailed model or adjustable internal parameters, sets RC apart from other machine learning paradigms, such as deep learning. 

While these systems have demonstrated impressive capabilities~\cite{jaeger2004,wyffels2010,pathak2017,lu2017,zimmermann2018}, classical RC has limitations, particularly in terms of interpretability and control. The internal state of the reservoir is often opaque, and perturbing or probing the system in a controlled manner can be challenging. These limitations have motivated the development of a new paradigm: \emph{next-generation reservoir computing (NGRC)}~\cite{NGRC2021}.

\begin{figure}[htp!]
\centering
    \begin{subfigure}[t]{\columnwidth}
        \centering
        \begin{tikzpicture}[>=Stealth, node distance=1.cm and 1.2cm]
            \tikzset{myarrow/.style={->, very thick}}
            \node (u0) {$\mathbf{u}(t)$};
            \node[right=of u0] (u1) {$\mathbf{u}(t+\tau)$};
            \node[left=of u0] (udotsL) {$\mathbf \cdots$};
            \node[right=of u1] (udotsR) {$\mathbf \cdots$};
            \node[below=of u0] (r0) {$\mathbf{r}(t)$};
            \node[below=of u1] (r1) {$\mathbf{r}(t+\tau)$};
            \node[left=of r0] (rdotsL) {$\mathbf \cdots$};
            \node[right=of r1] (rdotsR) {$\mathbf \cdots$};
            \coordinate (merge00) at ($(rdotsL)!0.8!(r0)$);
            \coordinate (merge0) at ($(r0)!0.7!(r1)$);
            \coordinate (merge1) at ($(r1)!0.85!(rdotsR)$);
            \draw[myarrow] (r0) -- node[left] {$\mathbf{W_{\rm out}}$} (u0);
            \draw[myarrow] (r1) -- node[left] {$\mathbf{W_{\rm out}}$} (u1);
            \draw[myarrow] (rdotsL) -- (r0);
            \draw[myarrow] (r0) -- (r1);
            \draw[myarrow] (r1) -- (rdotsR);
            \draw[myarrow] (rdotsL) -- node[below left] {} (r0);
            \draw[myarrow] (udotsL) to[out=-70, in=180] node[xshift=-9pt] {$\ff$} (merge00);
            \draw[myarrow] (u0) to[out=-70, in=180] node[xshift=-9pt] {$\ff$} (merge0);
            \draw[myarrow] (u1) to[out=-70, in=180] node[xshift=-9pt] {$\ff$} (merge1);
        \end{tikzpicture}
            \caption{\label{fig:map_diagram_a}Traditional reservoir computing according to Lu {\it et al.}~\cite{Ott2018} The reservoir dynamics at time $t+\tau$  is predicted via the reservoir function $f$, depending on the reservoir input $ \mathbf{u}(t)$ and the reservoir state $\mathbf {r}(t)$ at time $t$.}
    \end{subfigure}
    \vspace{0.5cm}
    \begin{subfigure}[t]{\columnwidth}
        \centering
        \begin{tikzpicture}[>=Stealth, node distance=1.cm and 1.2cm]
            \node (u1)  {$\mathbf u(t)$};
            \node [ left=of u1]  (u0)  {};
            \node [right=of u1]  (u2)  {$\mathbf u(t+\tau)$};
            \node [right=of u2]  (u3) {$\cdots$};
            \node [ below=of u0]  (d0)  {};
            \node [ below=of u1]  (d1)  {};
            \node [ below=of u2]  (d2)  {};
            \node [ below=of u3]  (d3)  {};
            \node [below=of u0] (uH0) {$\cdots$};
            \node [right=of d1, xshift=-20pt] (uH1)    {$\mathbf u_H(t)$};
            \node [right=of d2, xshift=-30pt] (uH2)    {$\mathbf u_H(t+\tau)$};
            \node [below=of u3] (uH3) {};
            \node [below=of d1] (r1) {$\mathbf r(t)$};
            \node [below=of d2] (r2) {$\mathbf r(t+\tau)$};
            \node [below=of d3] (r3) {$\cdots$};
            \draw[->, line width=1.2pt] (uH0) -- node[above,xshift =-10pt,yshift=-10pt] {$P$} (r1);
            \draw[->, line width=1.2pt] (r1) -- node[above,xshift =-15pt,yshift=0pt] {$W_{\rm out}$} (u1);
            \draw[->, line width=1.2pt] (u1) -- node[above, xshift=-4pt] {} (uH1);
            \draw[->, line width=1.2pt] (uH1) -- node[above, xshift=-10pt,yshift=-10pt] {$P$} (r2);
            \draw[->, line width=1.2pt] (r2) -- node[above,xshift =-15pt,yshift=0pt] {$W_{\rm out}$} (u2);
            \draw[->, line width=1.2pt] (u2) -- node[above, xshift=-4pt] {} (uH2);
            \draw[->, line width=1.2pt] (uH2) -- node[above, xshift=-10pt,yshift=-10pt] {$P$} (r3);
\end{tikzpicture}
        \caption{\label{fig:map_diagram_b}Next Generation Reservoir Computing  (NGRC) approach~\cite{NGRC2021} used in this article.}
\end{subfigure}
    \caption{\label{fig:map_diagram}
Comparison of traditional Reservoir Computing (RC)~\cite{Ott2018} and Next Generation Reservoir Computing (NGRC)~\cite{NGRC2021} schemes.}
    \label{fig:main}
\end{figure}
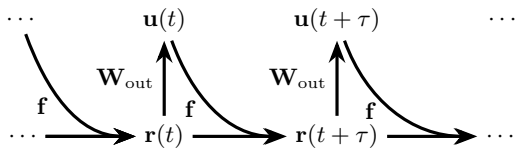
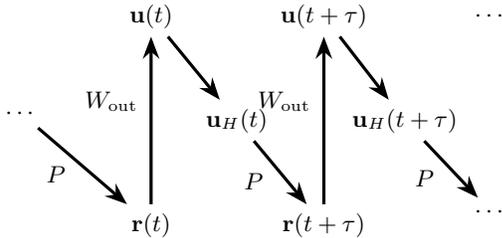

NGRC departs from the traditional architecture by eliminating the \emph{explicit} reservoir altogether. Instead, it constructs a high-dimensional feature space of dimension, $M$,  (corresponding to an \emph{implicit} reservoir) through a deterministic projection of the input signal. This results in a simple, fully transparent mapping from one time step to the next, where the system state is entirely known and controllable. Letting $u(t)$ denote the input and $r(t)$ the projected state, the evolution becomes a direct function $r(t+1)=F(r(t),u(t))$, enabling precise interventions and measurements. This is particularly advantageous in applications requiring system perturbation and response analysis, such as the computation of phase response curves~\cite{smeal2010phase,wedgwood2013phase,pietras2021reduced,cestnik_rosenblum_2018,cestnik_mau_rosenblum_2022}. See Sec.~\ref{sec:review} and Fig.~\ref{fig:map_diagram} for a more detailed comparison of the two methods.

When using NGRC a challenge remains to design a feature projection algorithm that is independent of the observation dimension, and the desired number of final features. 
For example, suppose the observation is $N$-dimensional with a time-embedding of $H$ steps. 
The dimension of the input to the reservoir is then $NH$. Further, suppose we use a feature expansion built on pair-wise combinations, we end up with $(NH)^2$ features. It is desirable that changing the observation dimension, $NH$, does not require changing the reservoir dimension $M$, i.e., the method is \emph{scalable}.

Our work adopts the next-generation reservoir computing framework with an emphasis on scalability and interpretability.
In particular, we introduce a pseudorandom nonlinear projection of time-delay embedded input as a simple and dimension-independent alternative to the polynomial-based projections used previously.
This projection maintains flexibility across different system sizes while avoiding the need to manage opaque internal reservoir states. The approach is illustrated on benchmark tasks such as attractor reconstruction, bifurcation diagram estimation, and phase recovery, showing stable long rollouts and robustness to partial or noisy data. 
We also include an exploratory demonstration of asymptotic phase estimation using NGRC.

The remainder of this paper is structured as follows. In Sec.~\ref{sec:review}, we briefly review classical RC and NGRC approaches. 
Sec.~\ref{sec:methods} explains our methods and  introduces the projection method used to lift the input into a high-dimensional space (Sec.~\ref{sec:projection}), followed by details on training the linear model (Sec.~\ref{sec:optimization}) and tracking prediction errors  (Sec.~\ref{sec:errors}). 
In Sec.~\ref{sec:examples} we present three example applications to demonstrate the method’s capabilities: the reconstruction of a chaotic attractor, the estimation of a bifurcation diagram, and the recovery of asymptotic phases of an oscillatory process.
Finally, we close with a discussion of the implications and limitations of our approach in Sec.~\ref{sec:discussion}.

\section{Brief review of reservoir computing\label{sec:review}}

We briefly review two reservoir computing approaches which form the conceptual foundation for the methods used throughout this paper:  the classical reservoir computing (RC) and the more recent next-generation reservoir computing (NGRC) scheme.

\subsubsection{Traditional Reservoir Computing (RC)}

We first review the classical reservoir computing (RC) framework (Lu {\it et al.}~\cite{Ott2018}), summarized diagrammatically in Fig.~\ref{fig:map_diagram_a} .  
Let $\uu(t)$ denote the input (or measurement) vector and $\rr(t)$ the internal state of the reservoir. The reservoir evolves according to the dynamical system,
\begin{equation}
    \dot{\rr}(t) = \ff(\rr(t), \uu(t)),
\end{equation}
where $\ff$ is an appropriately chosen function that is differentiable and typically high-dimensional.

A common choice for the \emph{reservoir function} $\ff$ is
\begin{equation}
    \ff = \gamma \left( -\rr + \tanh(\MM \cdot \rr + \sigma\, \WWi \cdot \uu) \right).
\end{equation}
Thus, updating the reservoir $\rr$ involves a nonlinear transformation (e.g., $\tanh$) of a linear combination of the reservoir state $\rr$ and the input signal $\uu$. The input signal, $\uu$, is linearly mapped into the reservoir via weights, $\WWi$, and the internal reservoir dynamics, $\rr$, is governed by a (possibly random) matrix, $\MM$. In practice, the exact form of $\MM$ and $\WWi$ does not need to be known, however, they need to remain unchanged throughout the phases of training and prediction.  

Reservoirs of this type are commonly referred to as ``echo state networks'' (ESNs)~\cite{HermansSchrauwen10,HermansSchrauwen11,GRIGORYEVA2018495,PhysRevE.103.062204,jaeger2004}. A key requirement is the \textit{echo state property}~\cite{Jaeger2001,ManjunathJaeger13,yildiz2012re}, which ensures that the reservoir state $\rr(t)$ in the asymptotic time limit does not depend on initial conditions but only on the input history. That is, the reservoir must effectively ``forget'' its initial state in order to respond reliably to the input.

The output of the reservoir, i.e., the estimated system state $\uu(t)$, is obtained as a readout from the reservoir state via a linear output layer,
\[
    \uu(t) = \WWo \cdot \rr(t),
\]
where $\WWo$ is trained, e.g., via linear regression, to learn the mapping $\WWo$ based on time series predictions. 
Often, the output layer uses a concatenation of $\rr(t)$ with its element-wise square, $\rr \odot \rr$, to break linear symmetries (e.g., sign reversals) and capture nonlinear structure; for further details on RC refer to, e.g., Ref.~\onlinecite{Ott2018}. 

In reservoir computing, the equations governing the reservoir dynamics and output are used during both the training and predicting phases. During the \emph{training phase}, the reservoir state $\rr(t)$ evolves according to the input signal $\uu(t)$, and the output $\uu(t)=\WWo\cdot\rr(t)$ is used for learning the mapping represented by the weight matrix $\WWo$. In the \emph{prediction phase}, the trained output layer $\WWo$ is applied to the reservoir state to determine the new input signal $\uu(t)$, which then in turn continues drive the evolution of the reservoir. (We note that some literature distinguishes variables for the training and prediction phases, but for our purposes we maintain the same notation for $\rr(t)$ and $\uu(t)$, regardless of whether they represent values for training or prediction phases.)
In this context, readout weights are parameters optimized during the training phase, whereas the projection dimension $M$ and other process defining quantities often are referred to as \emph{hyperparameters}, and are set before training and held fixed during prediction.

One of the main advantages of RC over traditional recurrent neural networks is that training is simple: only the output layer $\WWo$ needs to be trained (learned), while the reservoir dynamics itself remains fixed and untrained. For the readout, linear mapping to be expressive enough, the reservoir dimension must be sufficiently high. In effect, RC transforms a low-dimensional nonlinear problem into a high-dimensional linear one.

Moreover, a notable strength of reservoir computing is that the reservoir function $\ff$ can be chosen quite freely --- it does not even need to be precisely known. This flexibility enables the use of physical systems as reservoirs, particularly in photonic or optical platforms, which offer high bandwidth and fast computation speeds~\cite{larger2012photonic}.

\subsubsection{Next Generation Reservoir Computing (NGRC)}

\begin{figure}[htbp]
    \centering
    \includegraphics[width=0.49\textwidth]{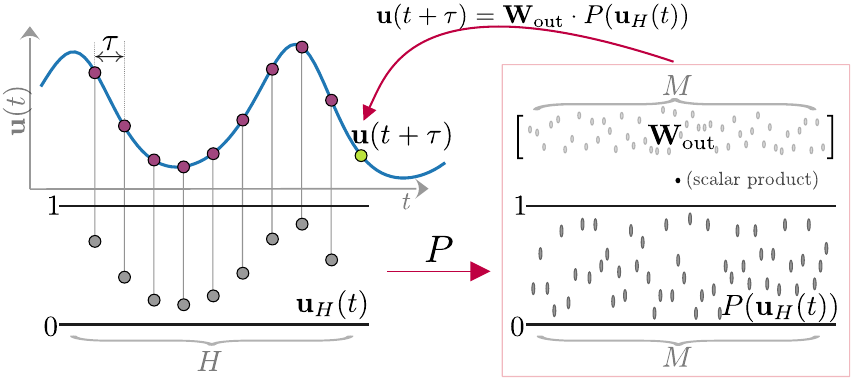}
    \caption{Schematic of the NGRC one-step predictor. A normalized, time-delay embedded state $\uu_H(t)$ is lifted by the nonlinear projection $P$ into a high-dimensional feature vector, which is mapped linearly by $\WWo$ to predict $\uu(t+\tau)$. The diagram shows the single scalar signal $\uu(t)$ case; for $N$ observables, $\uu_H$ has dimension $NH$ and $\WWo$ is an $M \times N$ matrix.  }
    \label{fig:ngrc_scheme}
\end{figure}

Next-generation reservoir computing (NGRC) is a simplified alternative to traditional RC. 
A schematic overview of the NGRC one-step prediction mechanism is shown in Fig.~\ref{fig:ngrc_scheme}. 
Instead of relying on an explicitly simulated dynamical system as the reservoir, NGRC uses a static, implicit reservoir structure based on nonlinear vector autoregression~\cite{NGRC2021}. The central idea is to take the (normalized) time-delay embedded state vector,
\[
\uu_H(t) := [\uu(t), \uu(t-\tau), \uu(t-2\tau), \dots, \uu(t-(H-1)\tau)],
\]
and map it nonlinearly into a high-dimensional space,
\[
\rr(t) = P(\uu_H(t)),
\]
as illustrated in Fig.~\ref{fig:map_diagram_b}.
The projected state $P(\uu_H(t))$ serves as the reservoir state $\rr(t)$, forming the implicit reservoir.
Prediction of the next state $\uu(t+\tau)$ is then performed --- as before in RC --- using a simple linear model,
\begin{equation}\label{eq:NGRC_scheme}
\uu(t+\tau) = \WWo \cdot P(\uu_H(t)),
\end{equation}
which constitutes the full one-step prediction mechanism of the NGRC approach.
The only trainable parameter is the output weight matrix $\WWo$ which can be learned by fitting the linear relation~\eqref{eq:NGRC_scheme} to a training data set.

This approach closely aligns with Koopman operator theory~\cite{koopman1931hamiltonian, mezic2005spectral, brunton2022modern}, which asserts that the dynamics of a nonlinear system can be represented by an infinite-dimensional linear operator acting on a suitably chosen set of observables.
In NGRC, the delay-embedded state vector $\uu_H(t)$ is mapped into a higher-dimensional space, resulting in the state $\rr(t)=P(\uu_H(t))$, which serves as such a set of observables, effectively lifting the system into a higher-dimensional function space where its evolution becomes approximately linear.
Rather than explicitly constructing the Koopman operator, NGRC learns a linear mapping $\WWo$ between $\rr(t)$ and the predicted next state $\uu(t+\tau)$. The linearity of $\WWo$ is consistent with the Koopman framework, but NGRC sidesteps the need to estimate the full operator. Finally, we note that, while this operation often is referred to as a ``projection'' in literature, we may more aptly think of this operation as a ``lifting'', as it maps the state into a higher-dimensional space rather than projecting onto a subspace.

An advantage of the NGRC architecture is that the time-delay embedding is explicit, eliminating the long transients required in traditional RC to forget initial conditions.
This simplification makes NGRC more practical for manipulating and initializing states, particularly in digital twin settings, where constructing meaningful initial conditions is often intractable due to the ``black box'' nature of the reservoir.

Despite its simplicity, NGRC has demonstrated strong performance on several challenging tasks~\cite{NGRC2021}, including (1) short-term forecasting, (2) attractor reconstruction, and (3) generalization to unseen regions of the state space.

\section{Methods\label{sec:methods}}

\subsection{Scope of This Work}

We introduce a simple, scalable projection scheme to construct the high-dimensional feature space used in next-generation reservoir computing. 
This projection involves pseudo-random mixing of the delay-embedded measurement vector $\uu_H$, and is detailed in Sec.~\ref{sec:projection}.

Using this projection, we train NGRC models on several representative dynamical systems, leveraging both complete and partial observations.
We demonstrate that these models accurately reproduce short-term trajectories, long-term attractors, asymptotic phases, and bifurcation diagrams.
Notably, we show that bifurcation diagrams can be inferred even when the bifurcation parameter varies in time as a correlated stochastic process
--- a scenario often arising in non-stationary data collection conditions, similar to the recurrent neural network approach in Ref.~\cite{cestnik_abel_2019}.

Our NGRC models remain stable over long prediction horizons and do not exhibit any transversal instabilities -- it seems that, even asymptotically, the integrator does not drift away and remains on the same attractor. 
This stability enables us to estimate bifurcation structure directly from a single long trajectory, employing a quasi-continuation approach.

\subsection{Projection $P$}\label{sec:projection}

In Ref.~\cite{NGRC2021}, the projection is achieved by constructing the Volterra series of the measurements $\uuH$, a series which contains monomials formed by multiplying the components of the input signal. In Ref.~\cite{ratas_pyragas_2024}, Chebyshev polynomials were used instead. 
Other NGRC lifting formulations, including kernel-based variants, have also been studied~\cite{grigoryeva_et_al_ortega_2025,gonon_et_al_ortega_2025}.
Various processes may thus be employed for this projection, but a limitation of such polynomial-based expansions is that the projection dimension grows with the size of the input vector, so increasing the observation dimension or embedding depth directly increases the number of features. To avoid this dependency, we adopt a \emph{pseudorandom projection scheme} based on pairwise mixing of input components. This approach makes the feature dimension $M$ a tunable parameter independent of the input size, thereby offering scalability across different system dimensions.

Our aim in introducing the pseudorandom projection is not to outperform existing feature constructions such as Chebyshev or Volterra expansions. 
Polynomial-based schemes are known to have near-optimal approximation properties for smooth functions on bounded intervals, so one should not expect a pseudorandom feature map to achieve lower approximation error for a given feature dimension $M$.
The motivation for our construction is different: it permits $M$ to be adjusted continuously and independently of the input dimension, without requiring discrete jumps between entire sets of basis functions
(e.g. choosing all triplets instead of all pairs sees the number of final features jump drastically from a square number to a cube number). 
This flexibility is often advantageous in practice, since any loss in approximation efficiency relative to polynomial bases can be offset by modestly increasing $M$.

The projection $P$ embeds (lifts) the state vector $\uuH$ into a higher-dimensional space, with $P(\uuH) = [p_1,p_2,p_3,...,p_M]$, where the dimension $M\gg 1$ is a free parameter, typically chosen to be sufficiently large.
For this task, we construct the projection using a pseudo-random shuffled sequence of indices, which is deterministic and reproducible.

First, we normalize the values of the input vector $\uu_H$ to lie within the unit interval $[0,1]$, as illustrated in Fig.~\ref{fig:ngrc_scheme}.
To maintain numerical stability, a small positive margin $\epsilon>0$ is introduced, ensuring the values fall within ($\epsilon,1-\epsilon$) and never leave the unit interval during computation.

The measurement vector $\uu$ is $N$-dimensional, and the time-delay embedded vector $\uuH$ contains $H$ delayed samples, resulting in a complete input vector of size $N\times H$ components:
\begin{align}
\begin{split}
[&u_1(t),u_1(t-\tau),\ldots,u_1(t-(H-1)\cdot\tau),\\
&u_2(t),u_2(t-\tau),\ldots,u_2(t-(H-1)\cdot\tau),\\
&\quad\quad\quad \vdots\\
&u_N(t),u_N(t-\tau),\ldots,u_N(t-(H-1)\cdot\tau)].
\end{split}
\label{eq:P_start}
\end{align} 
These initial values form the starting point for constructing the projected vector  $P(\uuH)$.
To distinguish them from the new features generated during the projection process, we assign non-positive indices:
\[
p_{-(N\cdot H-1)} ,\ p_{-(N\cdot H-2)},\ \ldots,\ p_{-1},\ p_0,
\]
where, for example, $p_{-(N\cdot H-1)} = u_1(t)$ and $p_0 = u_N(t - (H-1)\cdot\tau)$. 

Next, we generate the projected vector of size $M$ (with $M \gg N \times H$), by iteratively creating  new values, i.e., incrementing $m=1,\ldots, M$, from component pairs $p_i$ and $p_j$ with $i, j < m$.
Each new component is defined by the nonlinear combination,
\begin{equation}
p_m = (1 - p_i)^{p_j}\,,
\label{eq:el_combine}
\end{equation}
where the indices $i$ and $j$ are chosen from a pseudo-random but deterministic sequence, seeded for reproducibility.
Any duplicate values are discarded to prevent degeneracy and to ensure that feature matrix remains full-rank during training.

To illustrate this process, the first new value, $p_1$, is created by combining two pseudo-randomly selected elements from the initial set of values with non-positive indices ($p_i, p_j$ where $i, j \leq 0$).
The next value, $p_2$, can now incorporate $p_1$ along with the original components, and this process continues.
Consequently, the pool of available elements grows with each step, rapidly increasing the diversity of possible pairings used in the projection.

Importantly, the map~\eqref{eq:el_combine} preserves the unit interval: if $p_i, p_j \in (0, 1)$, then $p_m$ will also remains within that range.
This process is repeated until $M$ distinct components are obtained. In practice, we exclude the original components~\eqref{eq:P_start} from the final projected vector, although this step is not strictly necessary.
The resulting $M$-dimensional vector $[p_1, p_2, \ldots, p_M]$ is then used as the projected state $P(\uu_H)$.

\begin{figure}[htbp]
    \centering
    \includegraphics[width=0.42\textwidth]{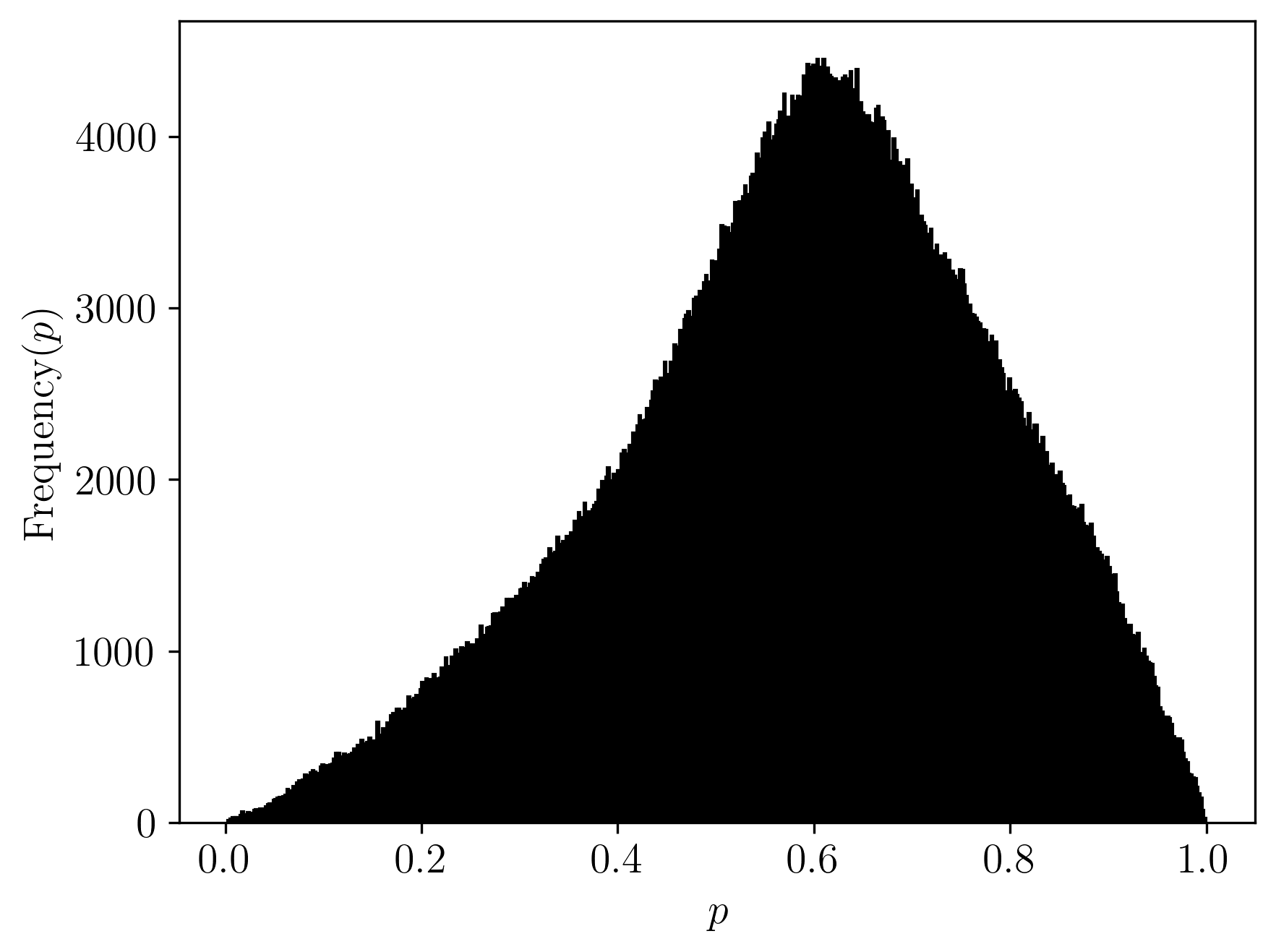}
    \caption{Distribution of $p_m$ values generated by the projection $P$ for a long time series of the Lorenz system~\eqref{lorenz}.
    The values remain bounded within the unit interval and are smoothly distributed, reflecting the stable and well-scaled nature of the projection scheme. }
    \label{fig:distr}
\end{figure}

A notable property of this projection scheme is that it preserves the unit interval: all projected components lie within $[0,1]$, are typically of the same order, and do not aggregate around particular values.
Figure~\ref{fig:distr} shows the distribution of $p_m$ values for a trajectory of the Lorenz system. 
The values are distributed smoothly and densely across the interval, contrasting monomial-based projections which often produce extreme values, either very small (near zero) or excessively large, depending on the scaling of the input signal.

\subsection{Optimizing the linear model}\label{sec:optimization}
Assuming that the projection $P$ has been chosen appropriately, we can expect that the linear relation~\eqref{eq:NGRC_scheme} provides a good approximation of the system dynamics and can be fitted directly to data.
Given the observed states \( \uu(t) \) at several time points, we construct their delay-embedded forms \( \uu_H(t) \), compute the corresponding projections \( P(\uu_H(t)) \), and then optimize the relation~\eqref{eq:NGRC_scheme} over all available training time points.
Fitting the model corresponds to finding the optimal output weight matrix \( \WWo \).  
Since the relation is linear, the optimization is straightforward and can be performed using standard techniques such as least squares regression or singular value decomposition~\cite{press2007numerical}.

The state vector \( \uu \) has dimension \( N \), while the projected vector \( P(\uu_H) \) is a vector of size $M$, where typically \( M \gg 1 \) and \( N \) may be small (even \( N = 1 \)).  
Thus, the output matrix \( \WWo \) has dimensions \( N \times M \).  
Each time point contributes one instance of the relation~\eqref{eq:NGRC_scheme}, and the entire training dataset over \( T \) time points can be expressed as:
\begin{subequations}
\begin{align}
\UU &= [\uu(t+\tau),\ \uu(t+2\tau),\ \uu(t+3\tau),\ \ldots,\ \uu(t+T\tau)] \\
\PP &= [P(\uu_H(t)),\ P(\uu_H(t+\tau)),\ \ldots,\ P(\uu_H(t + (T-1)\tau))]
\end{align}
\end{subequations}
which yields the matrix relation:
\begin{equation}
\UU = \WWo \PP.
\end{equation}
The optimal output weights \( \WWo \) are then obtained via least squares minimization:
\begin{equation}\label{eq:Wsolution}
\WWo = \UU \PP^T \left( \PP \PP^T \right)^{-1}.
\end{equation}
In practice, it is common to add Tikhonov regularization to improve conditioning and suppress overfitting:
\begin{equation}
\WWo = \UU \PP^T \left( \PP \PP^T + \lambda I \right)^{-1},
\end{equation}
where $\lambda > 0$ is a regularization parameter and $I$ the identity matrix.

\subsection{Tracking errors}\label{sec:errors}
Once the model \( \WWo \) has been determined, we evaluate its accuracy by tracking the mean squared error of the one-step prediction. For each time point $n$, we compare the true next state $\uu((n+1)\tau)$ with the predicted state $\WWo P(\uu_H(n\tau))$, where $P(\uu_H(n\tau))$ is the projected feature vector constructed from the previous $H$ delay-embedded states. The mean squared error is then
\begin{equation}
E = \frac{1}{T} \sum\limits_{n=1}^T \left\| \uu((n+1)\tau) - \WWo\, P(\uu_H(n\tau)) \right\|^2,
\label{sqerr}
\end{equation}
where $||\cdot||$ denotes the Euclidean norm and \( T \) is the number of prediction steps.
This error can be computed both on the training dataset and on a separate validation set that was not used during fitting.

\subsection{Validation}

Validation is the assessment of model performance on data not used during training, providing a measure of how well the model generalizes to new, unseen inputs.

The projection dimension \( M \) is a free parameter and can be treated as a hyperparameter of the method.
While the training error~\eqref{sqerr} typically decreases with increasing \( M \), monitoring the validation error is essential to avoid overfitting (i.e., as $M$ becomes larger, the model can fit the data, including fluctuations or noise, too closely, effectively ``remembering'' and thus performs poorly on unseen data in the validation set).

Additionally, the least squares solution relies on the invertibility of the matrix \( \PP \PP^T \), which can become problematic for simple systems or when the projection dimension \( M \) is large. 
A practical way to mitigate this issue is to add a small amount of measurement noise to the original state vectors \( \uu \), which improves numerical stability and can also act as a form of regularization. 
Conditioning issues and the sensitivity of NGRC regression to the lifted feature representation have been analyzed in Ref.~\cite{zhang2025chaos}, showing how noise and regularization can improve numerical stability. 

\begin{figure}[htbp]
    \centering
    \begin{overpic}[width=0.4\textwidth]{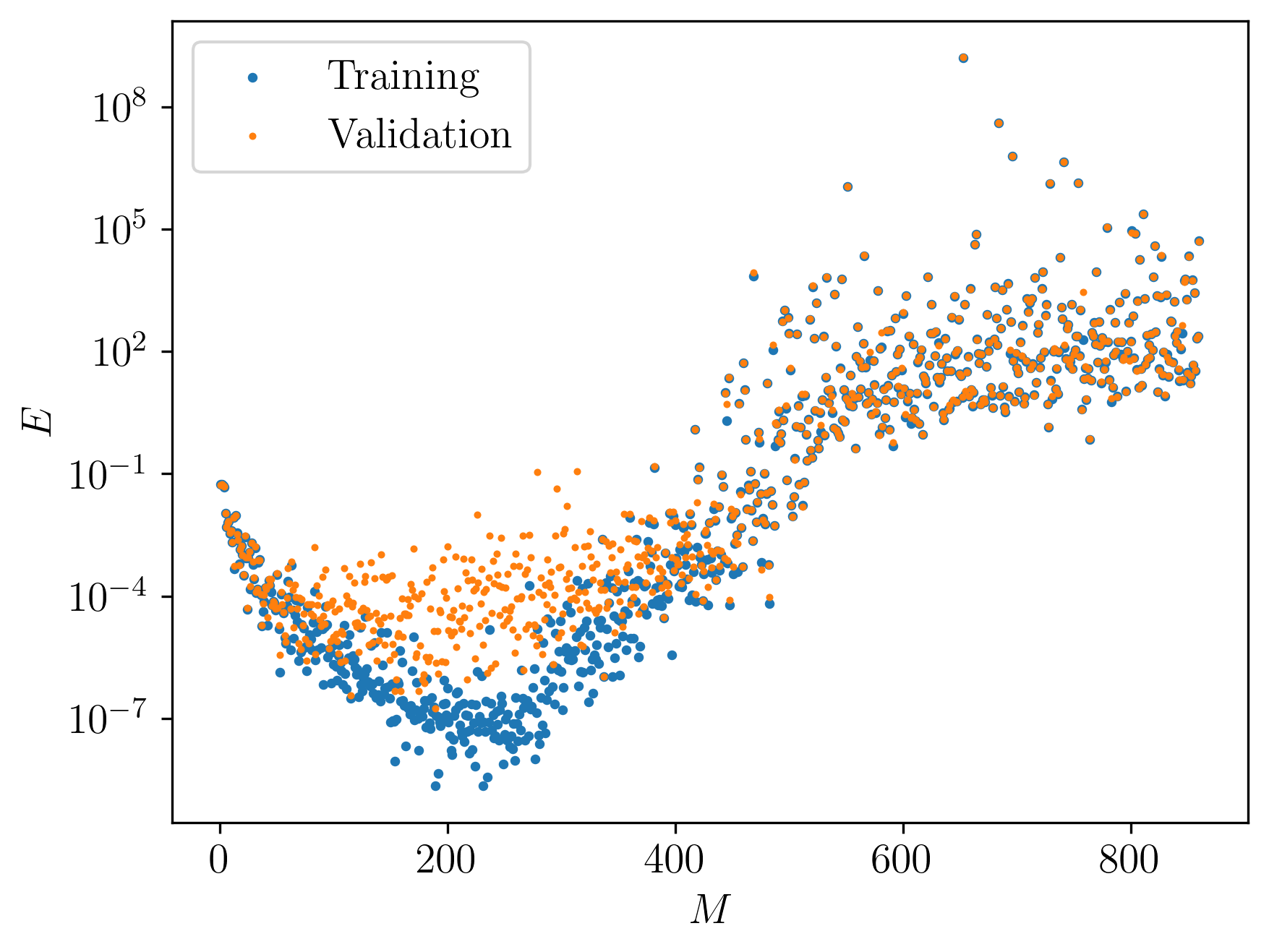}
    \put(15,13){$(a)$}
    \put(44,68.5){no noise}
    \put(44,63.5){no regularization}
	\end{overpic}
	\begin{overpic}[width=0.4\textwidth]{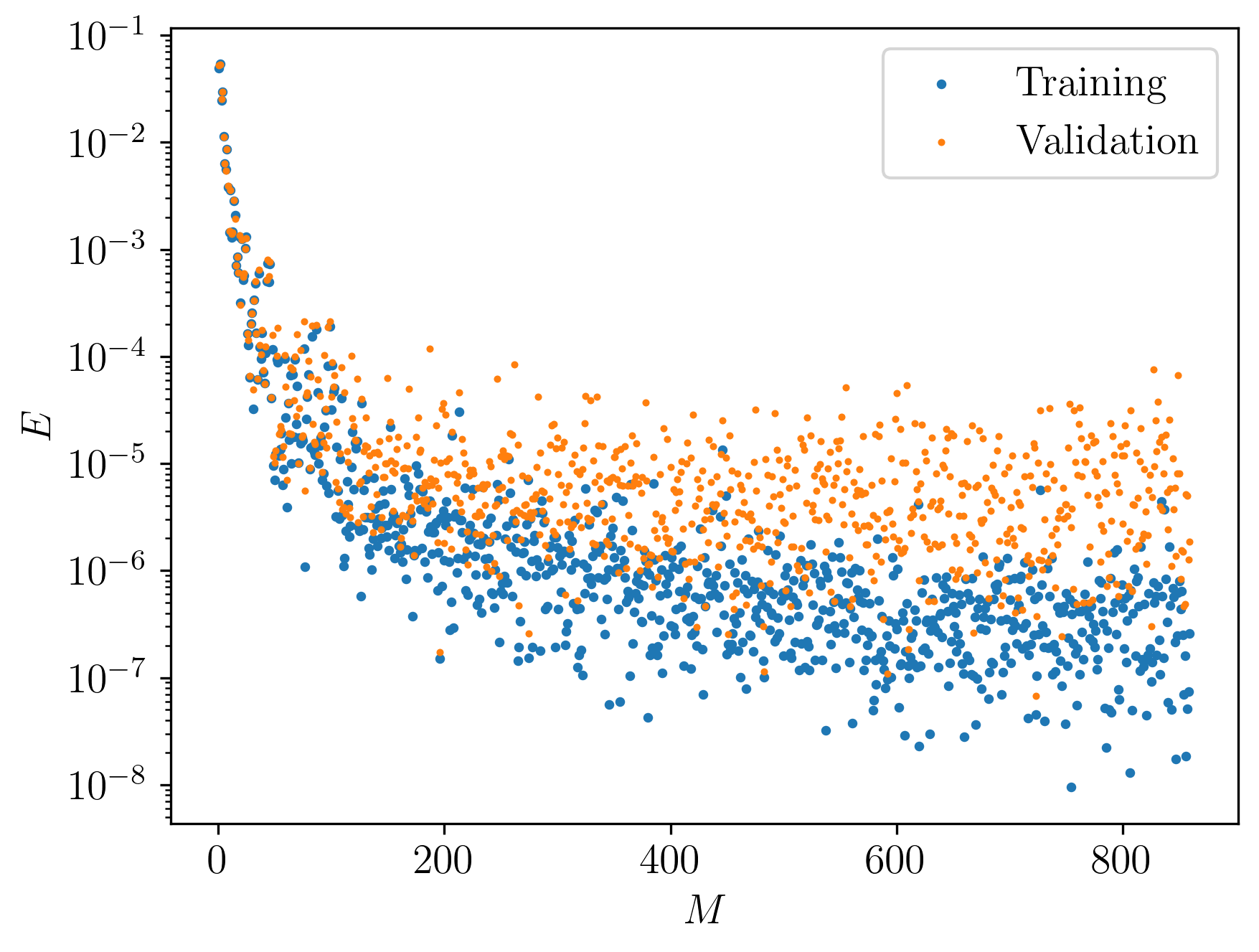}
	\put(15,13){$(b)$}
    \put(21,68.5){Tikhonov regularization}
    \put(21,63.5){$\lambda=0.01$ (no noise)}
	\end{overpic}
	\begin{overpic}[width=0.4\textwidth]{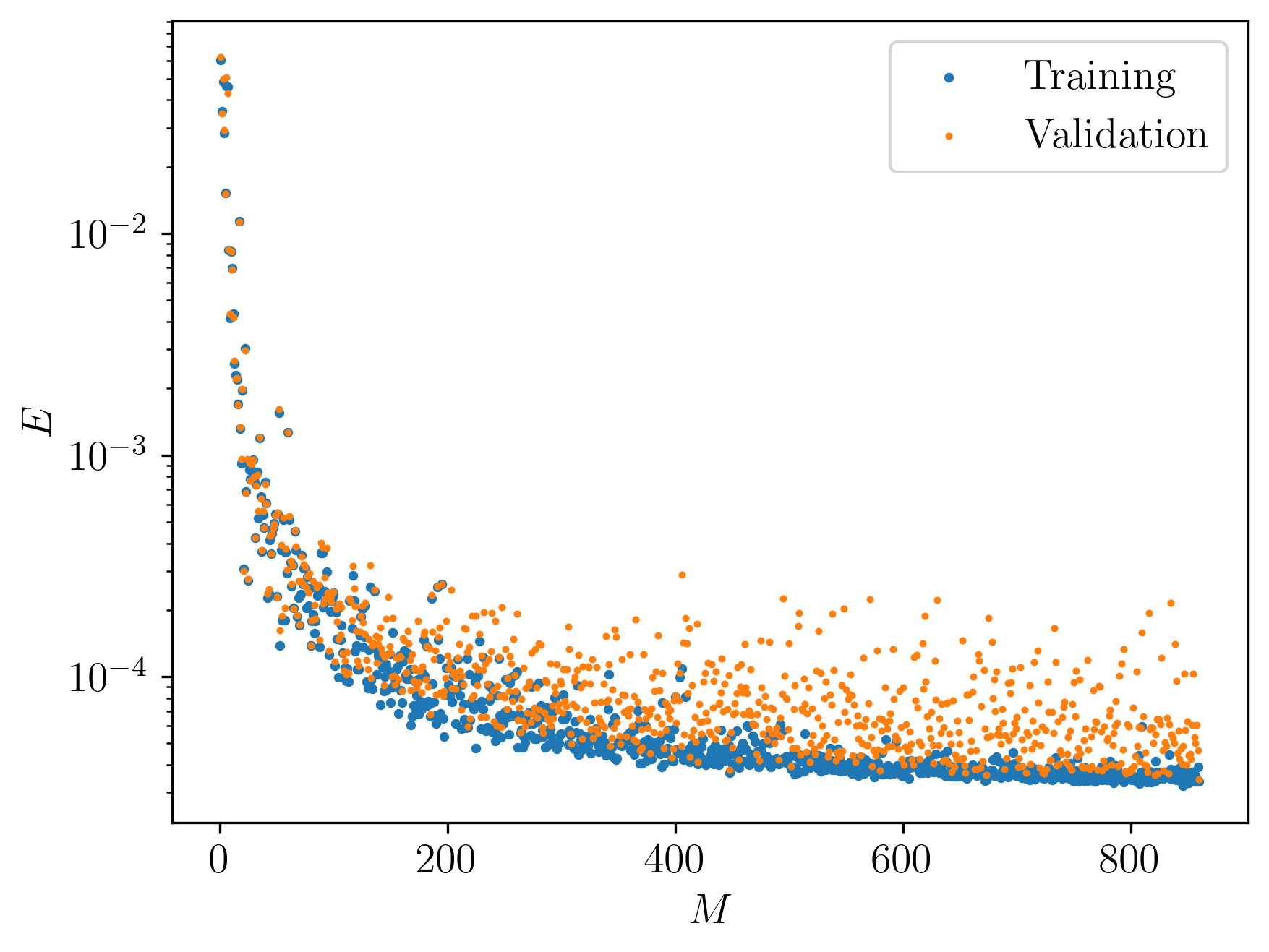}
	\put(15,13){$(c)$}
    \put(21,68.5){1\% measurement noise}
    \put(21,63.5){(no Tikhonov reg.)}
	\end{overpic}
    \caption{Mean-squared one-step prediction error as a function of the projection dimension $M$ under three  regularization scenarios: 
    $(a)$ no regularization, $(b)$ Tikhonov regularization with $\lambda = 0.01$, and $(c)$ addition of 1\% Gaussian measurement noise to the training inputs. 
    $(a)$ Without regularization, errors initially decrease but instabilities appear for larger $M$. 
    $(b)$ Tikhonov regularization suppresses high-$M$ instabilities, but mild overfitting persists, with training errors consistently below validation errors. 
    $(c)$ Adding measurement noise acts as an implicit regularizer by broadening the sampled state space, yielding training and validation errors that remain comparable across the full range of $M$. 
    This motivates the use of 1\% measurement noise in subsequent experiments as a practical balance between stability and predictive accuracy. Excessive noise, however, degrades performance, so a trade-off applies. All results use long trajectories ($T = 10^5$) for both training and validation.}
	\label{figerr}
\end{figure}

Figure~\ref{figerr} shows the training and validation errors~\eqref{sqerr} as a function of \( M \), for two different levels of Gaussian measurement noise.  
In the noise-free case, the error decreases only up to about \( M = 200 \), after which it increases sharply, leading to models that are effectively unusable.  
In contrast, adding 1\% Gaussian noise stabilizes the inference and prevents this error blow-up for moderate values of \( M \).  
However, excessive noise can also degrade model quality, so we selected the 1\% level as a reasonable compromise --- at which point the training and validation errors remain comparable across a range of \( M \). We consistently use 1\% measurement noise for all examples going further. 

\subsection{Long-term stability}
While one-step prediction accuracy is informative, it does not guarantee reliable long-term behavior when the predictor is iterated autonomously. 
Transverse instability has been identified as an important limitation for NGRC-type models in long-term autonomous prediction~\cite{ratas_pyragas_2024}, 
and recent work has examined both intrinsic stability conditions~\cite{santos2025arxiv} and complementary post-processing strategies for enhancing long-horizon robustness~\cite{berry_das_2025}. 
Similar stability concerns have also been reported in classical RC, where training noise has been shown to act as a regularizer that improves long-horizon autonomous behavior~\cite{wikner_et_al_ott_2024}. 
We therefore assess long-term stability under three regularization settings: (a) Tikhonov regularization with $\lambda=0.01$ and no measurement noise, (b) the same Tikhonov regularization with $0.1\%$ measurement noise added to the training inputs, and (c) $1\%$ measurement noise without Tikhonov regularization. For each setting we train 100 independent NGRC models and, for each model, generate 10 autonomous prediction trajectories from slightly perturbed initial conditions.

\begin{figure}[htbp]
    \centering
    \begin{overpic}[width=0.49\textwidth]{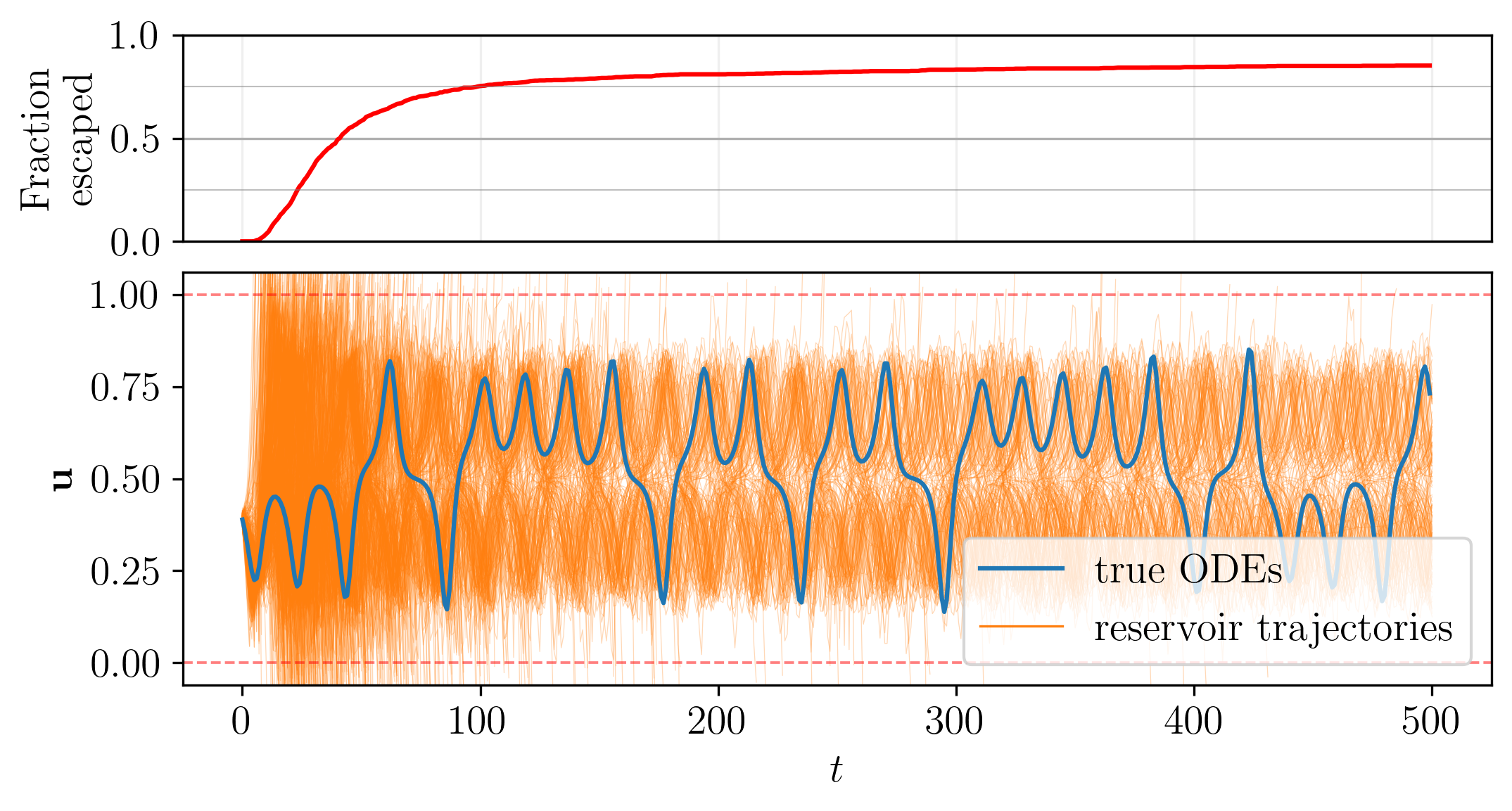}
    \put(13,47){$(a)$}
    \put(43,42){Tikhonov regularization $\lambda=0.01$}
    \put(43,39){0 measurement noise}
	\end{overpic}
	\begin{overpic}[width=0.49\textwidth]{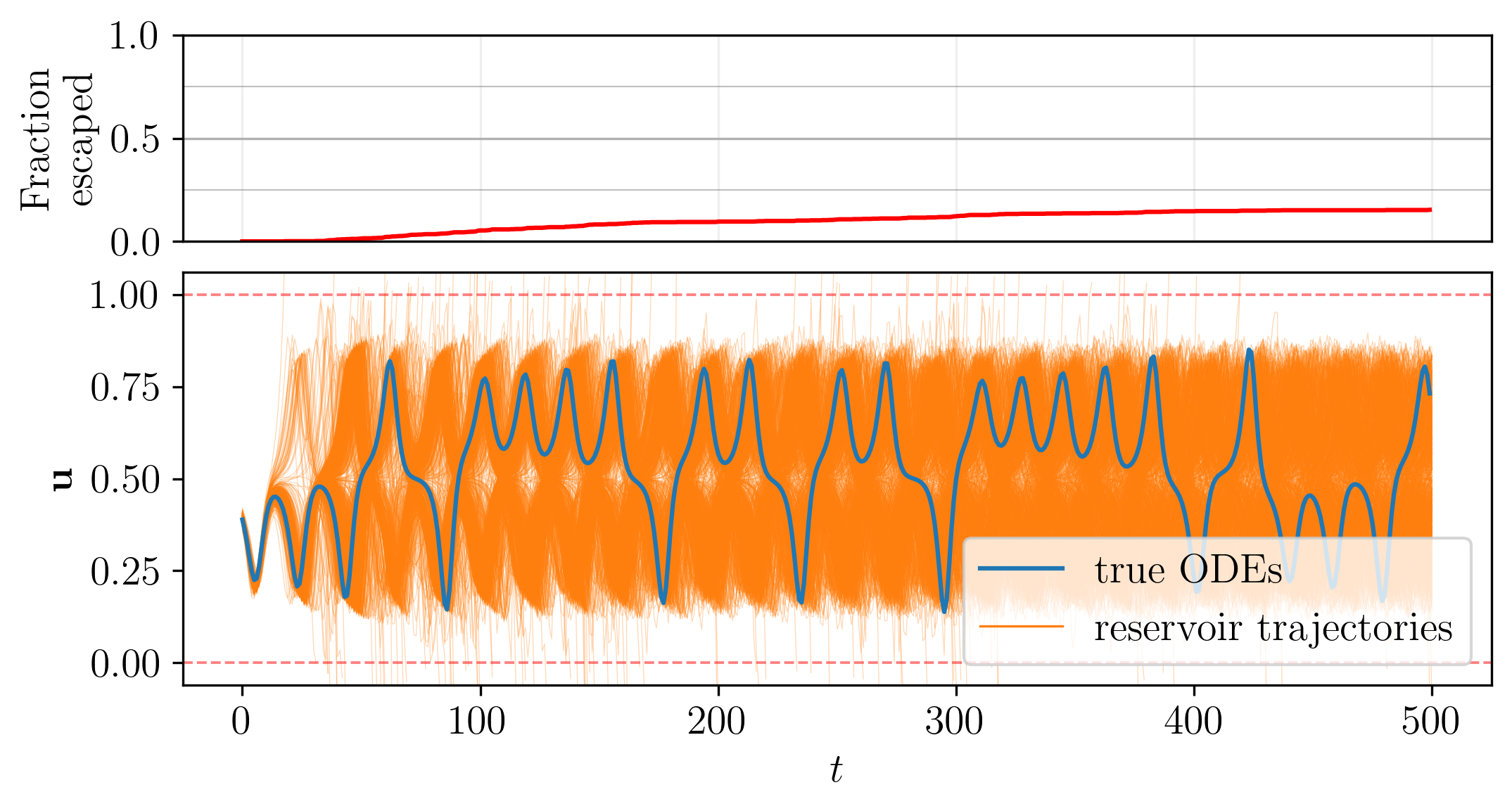}
	\put(13,47){$(b)$}
    \put(43,47){Tikhonov regularization $\lambda=0.01$}
    \put(43,44){$0.1\%$ measurement noise}
	\end{overpic}
	\begin{overpic}[width=0.49\textwidth]{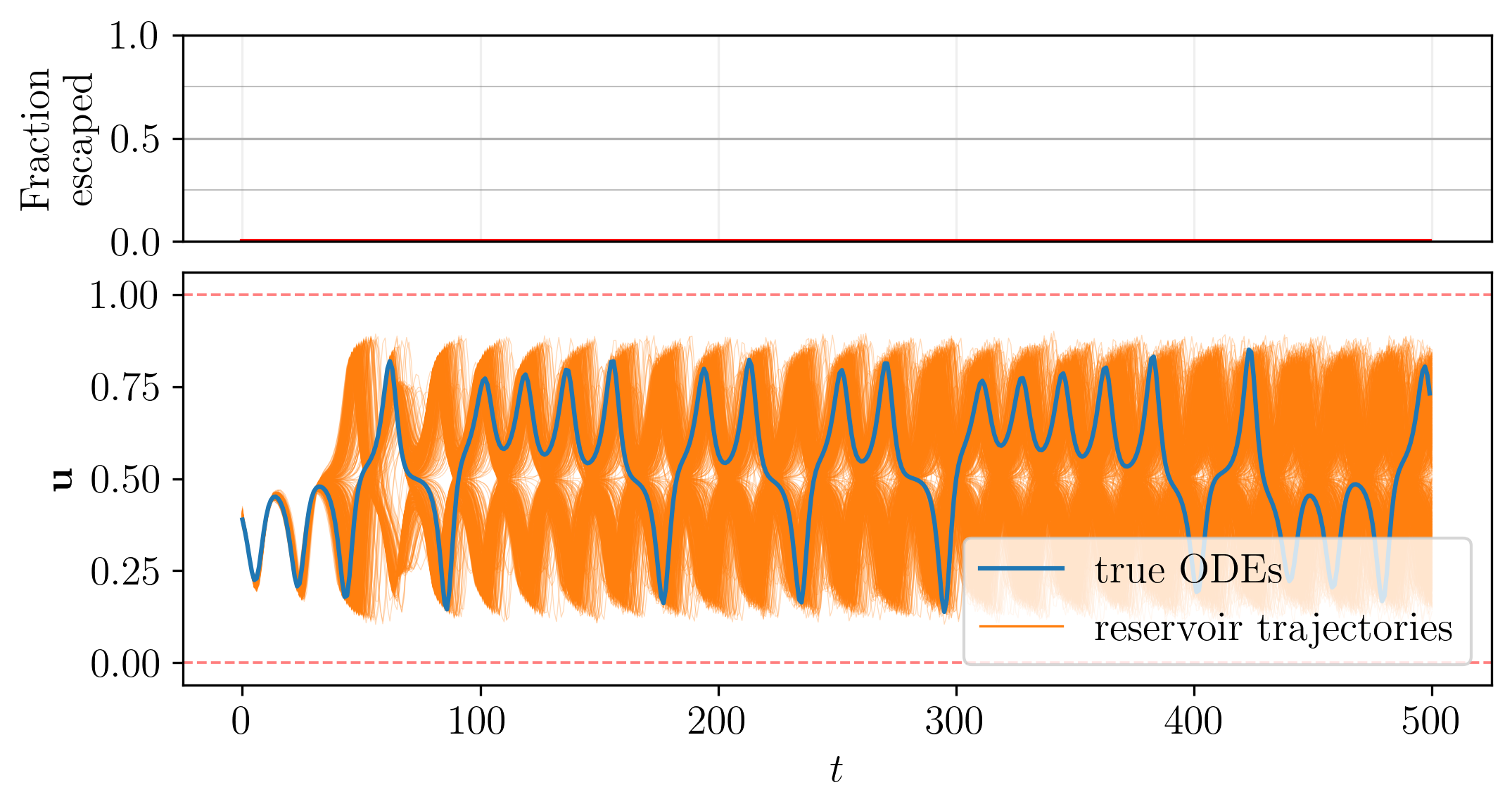}
	\put(13,47){$(c)$}
    \put(45,47){$1\%$ measurement noise}
    \put(45,44){(no Tikhonov regularization)}
	\end{overpic}
    \caption{Long-term autonomous prediction test for NGRC under three regularization settings: 
    $(a)$ Tikhonov regularization with $\lambda=0.01$ and no measurement noise, 
    $(b)$ the same Tikhonov regularization with $0.1\%$ measurement noise added to the training inputs, and 
    $(c)$ $1\%$ measurement noise without Tikhonov regularization. For each setting, 100 independently trained models are rolled out from 10 slightly perturbed initial conditions, and trajectories are compared with a reference solution of the true ODE system~\eqref{lorenz}. Because all signals are normalized to $[0,1]$, a trajectory is considered divergent once it leaves this interval, at which point the rollout is terminated. 
    The fraction of runs that diverged is shown alongside the trajectories. 
    $(a)$ sees rapid transverse instability and early divergence in most runs; 
    $(b)$ shows substantially reduced escape rates; 
    $(c)$ demonstrates the strongest stability, with no trajectories leaving escaping over the displayed time window. 
}
	\label{fig:longterm}
\end{figure}

Figure~\ref{fig:longterm} shows representative prediction trajectories together with a reference trajectory from the true ODE system (here we used Lorenz~\eqref{lorenz}). Because all signals are normalized to lie in $[0,1]$, we consider a trajectory to have diverged once it leaves this interval and terminate its evolution at that point. The fraction of escaped trajectories is plotted alongside the trajectories.

In case (a), most trajectories exhibit rapid transverse instability and leave the admissible range after only a few steps. In case (b), even a very small amount of measurement noise noticeably improves stability, reducing the escape rate and allowing trajectories to remain near the attractor for longer times. In case (c), training with $1\%$ measurement noise yields the most robust behavior: over the time window tested, none of the trajectories escapes the interval, despite the absence of Tikhonov regularization. This suggests that measurement noise acts as an effective implicit regularizer by broadening the sampled state space and suppressing fragile directions. For this reason, we adopt $1\%$ measurement noise in all subsequent experiments.

\section{Examples\label{sec:examples}}

\subsection{Reconstruction of attractor from partial observations}\label{sec:ex1}
We train the reservoir model~\eqref{eq:NGRC_scheme} using the random projection $P$ described in Sec.\ref{sec:projection}. The training data comes from two standard chaotic systems, the Lorenz system  and the Rössler system, which are three dimensional models formulated in variables $x(t),y(t),z(t)$ see Eqs.~\eqref{lorenz} and Eqs.~\eqref{rossler}, respectively. We generate long trajectories from each system, but we only use the scalar variable $x(t)$ for training. The model is trained for one-step prediction. After training, it can be used to predict long-term evolution by iterating its own outputs.

First, we compare short-time predictions from the reservoir model with the true trajectories. Next, we let the model run freely for a long time and plot the time-embedded trajectory alongside one from the true system. 

For the experiments, we fix some (hyper-)parameters: $M = 1000$ reservoir nodes, 1\% measurement noise, and a timeseries length $T = 10^5$. We observe only one variable ($x$) from each system, so $N = 1$.

\subsubsection{Lorenz system}
We consider the equations
\begin{equation}
\begin{aligned}
\dot{x} &= \sigma(y-x)\\
\dot{y} &= x(\rho-z)-y\\
\dot{z} &= xy-\beta z
\end{aligned}
\label{lorenz}
\end{equation}
with the standard parameters $\sigma = 10, \rho = 28, \beta = 8/3$~\cite{lorenz1963deterministic}. 

\begin{figure}[htbp]
    \centering
    \includegraphics[width=0.4\textwidth]{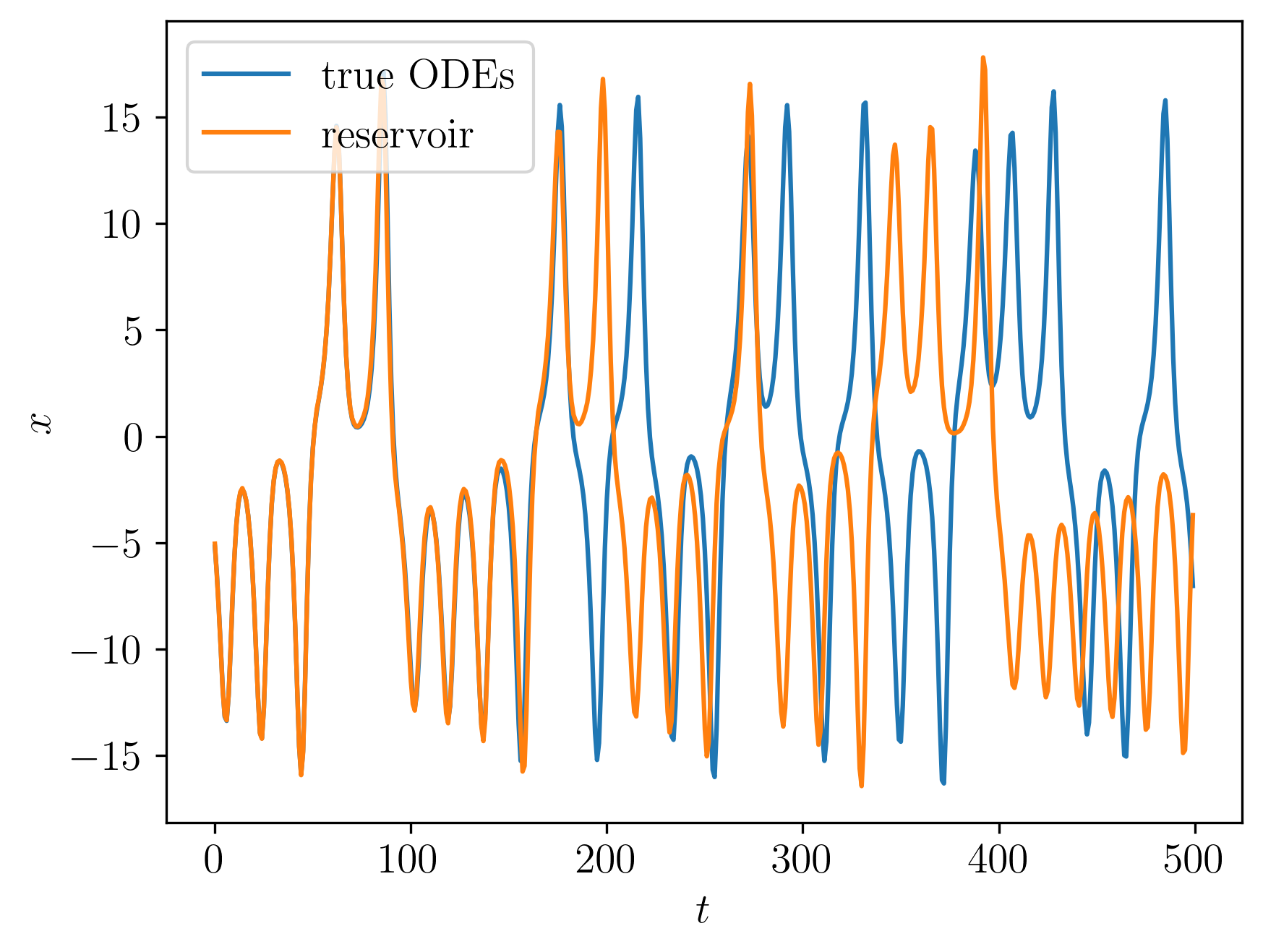}
    \caption{Short-term trajectories of the Lorenz system. The true trajectory is shown in blue, and the reservoir model prediction in orange. The model was trained using only the scalar variable $x(t)$. Time axis is units of time-sampling step $\tau = 0.04$. }
    \label{figtraj}
\end{figure}

Figure~\ref{figtraj} shows the true Lorenz trajectory (test data) alongside the trajectory generated by the reservoir model using iterative one-step predictions. As expected for a chaotic system, the two trajectories diverge from each other after some time, where small differences amplify exponentially in time. This divergence occurs on a timescale comparable to the Lyapunov time due to the finite accuracy of our one-step predictor.

\begin{figure}[htbp]
    \centering
    \includegraphics[width=0.4\textwidth]{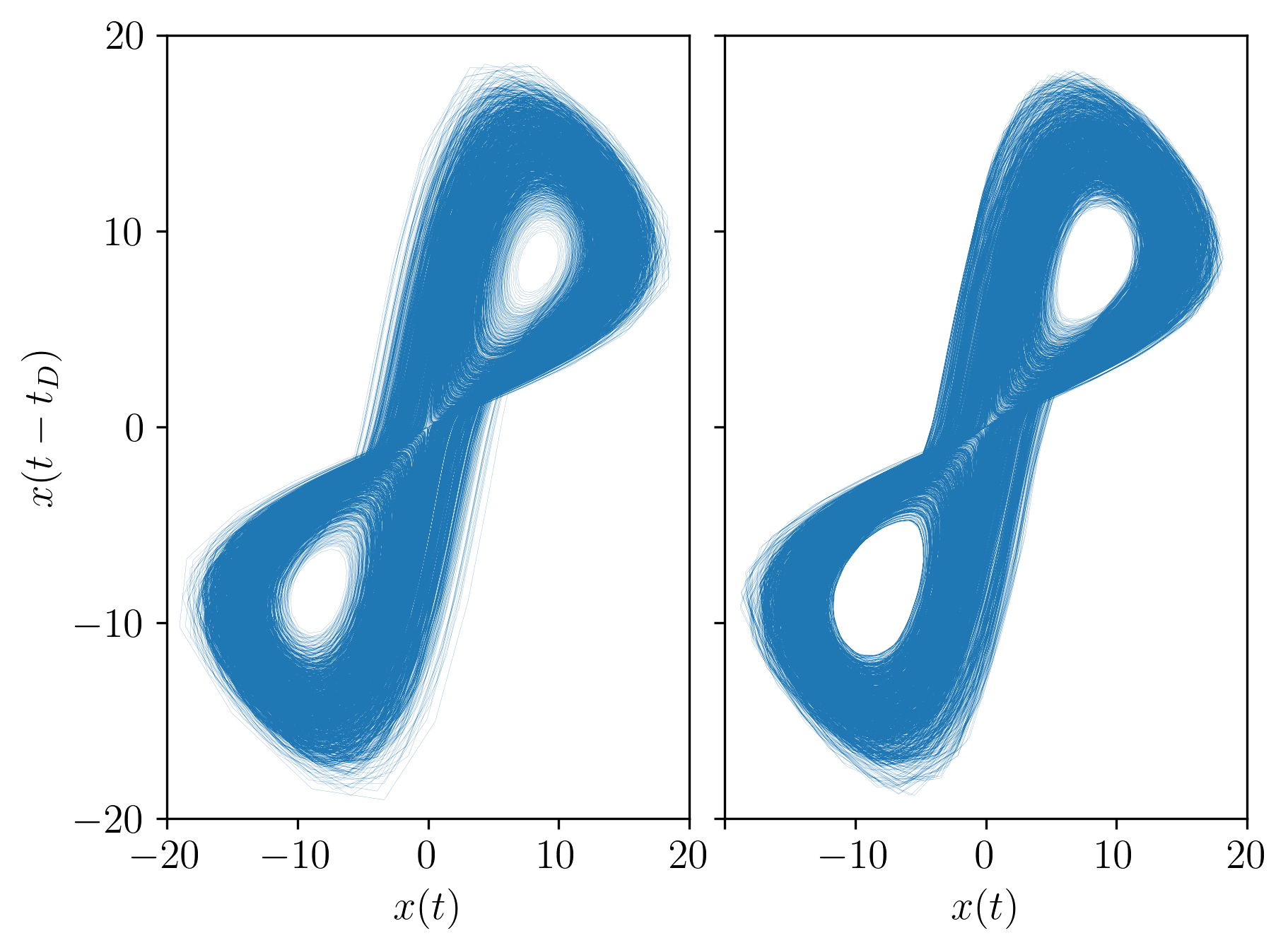}
    \caption{Delay-embedded Lorenz attractor reconstruction. The true attractor is shown on the left, and the reconstructed attractor on the right. The model was trained using only the scalar variable $x(t)$. Time delay for plot is $3\tau = 0.12$. }
    \label{figattr}
\end{figure}

Figure~\ref{figattr} compares the true Lorenz attractor obtained via numerical integration of Eqs.~\eqref{lorenz} with the one reconstructed from the reservoir model using delay embedding. The reconstruction produces a good match.

\subsubsection{Rössler system}
We consider the equations
\begin{equation}
\begin{aligned}
\dot{x} &= -y-z\\
\dot{y} &= x+ay\\
\dot{z} &= b+z(x-c)
\end{aligned}
\label{rossler}
\end{equation}
with parameters $a = 0.2, b = 0.4, c = 5.7$~\cite{roessler1976equation}. 

\begin{figure}[htbp]
    \centering
    \includegraphics[width=0.4\textwidth]{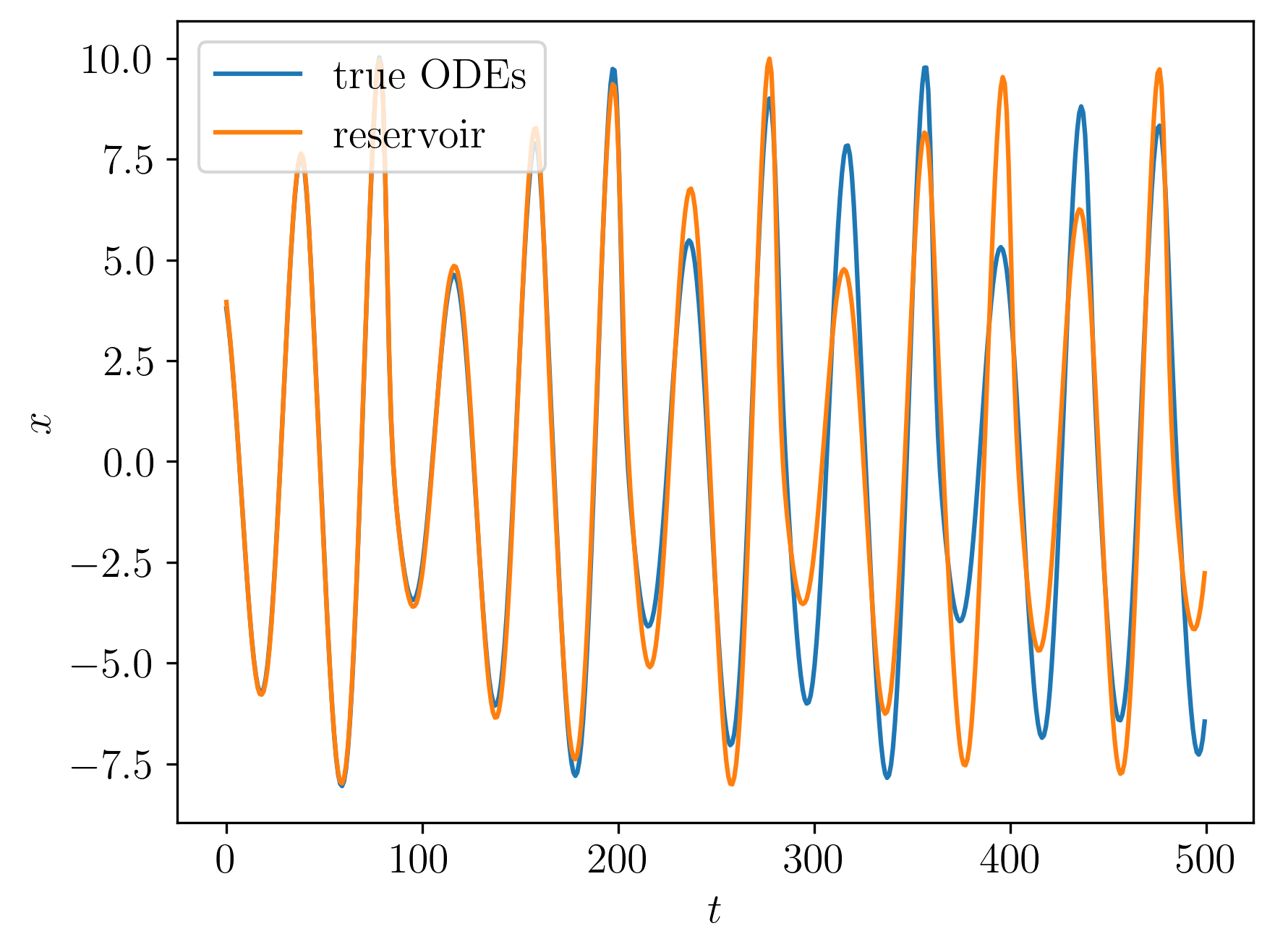}
    \caption{Short-term trajectories of the Rössler system. The true trajectory is shown in blue, and the reservoir model prediction in orange. The model was trained using only the scalar variable $x(t)$. Time axis is units of time-sampling step $\tau = 0.15$.}
    \label{figtraj_ros}
\end{figure}

Figure~\ref{figtraj_ros} shows the true Rössler trajectory, obtained by numerically integrating Eqs.~\eqref{rossler}, and the one produced by the reservoir model through iterative one-step predictions. As with the Lorenz case, the trajectories diverge after some time, consistent with the system's chaotic nature.

\begin{figure}[htbp]
    \centering
    \includegraphics[width=0.4\textwidth]{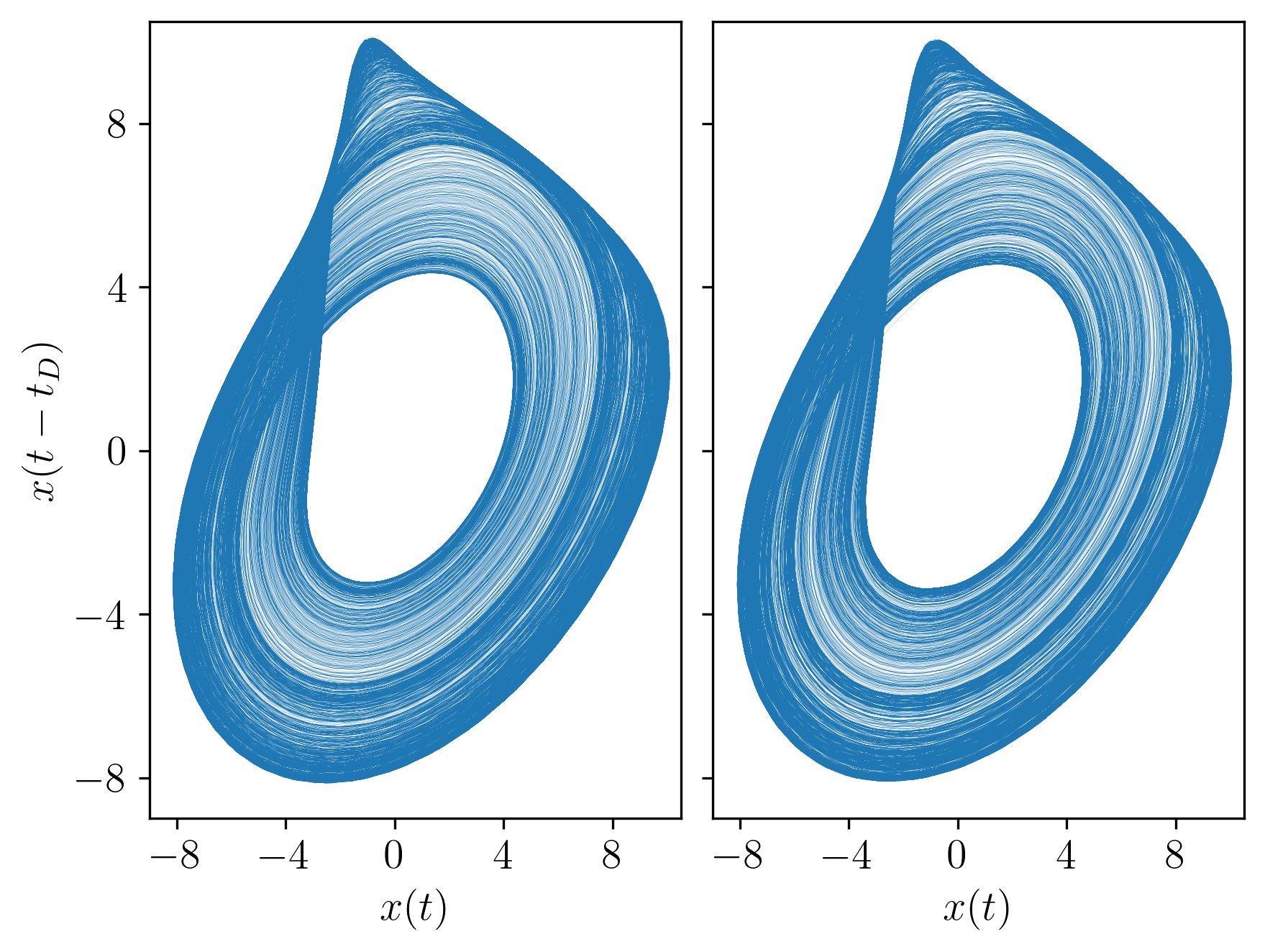}
    \caption{Delay-embedded Rössler attractor reconstruction. The true attractor is shown on the left, and the reconstructed attractor on the right. The model was trained using only the scalar variable $x(t)$. Time delay for plot is $8\tau = 1.2$. }
    \label{figattr_ros}
\end{figure}

Figure~\ref{figattr_ros} compares the true Rössler attractor with the attractor reconstructed from the reservoir model using delay embedding. Again, the original and reconstructed attractors match very well.

\subsection{Reconstruction of bifurcation diagram}\label{sec:bifurcation}\label{sec:ex2}
We consider the Rössler system~\eqref{rossler} and add a variable control parameter $\eta$ to the right-hand-side of the $y$ equation. 
The reservoir then observes the resulting system states and learns to reproduce the system's behavior for different $\eta$. 
The variation of $\eta$ allows us to probe the bifurcation structure of the system. 
In the following, we first use several data sections corresponding to different constant $\eta$ values to reconstruct the bifurcation diagram, 
and secondly, consider a time-dependent, stochastic $\eta(t)$ to mimic less controlled real world situations.

\begin{figure}[htbp]
    \centering
    \includegraphics[width=0.42\textwidth]{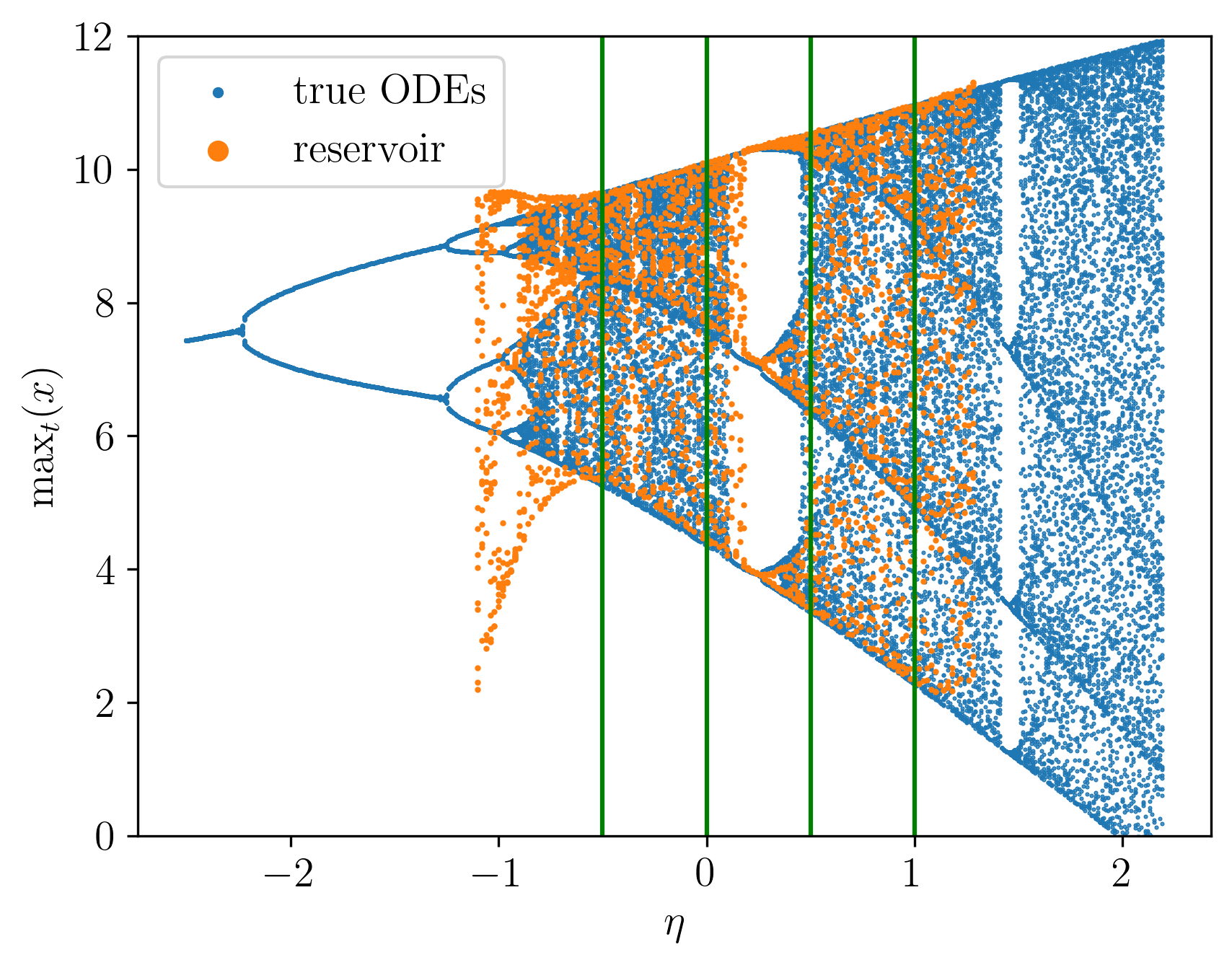}
    \caption{Bifurcation diagram of the true Rössler system~\eqref{rossler} (blue) and the reconstruction from the reservoir (orange). Green vertical lines mark the four input parameter values $\eta$ used for training. The diagram is computed by plotting the local maxima of $x(t)$. Training used long trajectories ($T = 10^5$), all three state variables ($x,y,z$), reservoir size $M = 6000$, delay dimension $H = 25$, and 1\% measurement noise.}
    \label{biffurcation}
\end{figure}

First, we generate trajectories for four different constant values of $\eta$ (indicated by vertical lines in Fig.~\ref{biffurcation}) and use them to train a single reservoir model.
After training, we estimate the bifurcation diagram by varying $\eta$ in the trained reservoir. The diagram is constructed by plotting the local maxima of the variable  $x(t)$ over time, see Fig.~\ref{biffurcation}. The diagram is obtained from only one reservoir trained on data corresponding to four values of $\eta$ as input to the reservoir.

The reconstruction is accurate near the trained parameter values, and is well interpolated between them. As $\eta$ moves farther away from $\eta$ values used for training, the reservoir-generated attractors increasingly deviate from the true system attractor. This is expected, since the model extrapolates beyond the regimes it has seen during training. 
Such deviations are a common limitation of data-driven models that rely on local interpolation rather than a global parameterization of the underlying dynamics. In particular, NGRC has no built-in mechanism to guarantee correct dynamics outside the region explored during training, and uncertainties can grow rapidly once the model leaves this domain. 
In Fig.~\ref{biffurcation}, this is most prominent near
$\eta=-1$, where the reservoir predicts qualitatively different behavior~\cite{confabulationreservoir2025}.

Next, we consider a more challenging and practically relevant scenario.
In many real-world applications, it is not possible to collect data under strictly controlled conditions where the control parameter, $\eta$, remains constant.
Instead, the parameter often varies in time and can be measured only as a noisy signal.

To mimic this situation, we drive the Rössler system ~\eqref{rossler} with a stochastic parameter $\eta(t)$ evolving according to an Ornstein-Uhlenbeck (OU) process~\cite{ornstein1930brownian}. 
The OU process is a standard way to model noise with a finite correlation time. It describes a fluctuating variable that is randomly driven but does not change abruptly, producing smoothly varying noise rather than instantaneous (white-noise) jumps. 
\begin{equation}
\begin{aligned}
\dot{x} &= -y-z\\
\dot{y} &= x+ay + \eta\\
\dot{z} &= b+z(x-c)\\
\dot{\eta} &= \frac{1}{\tau_c}\left( -\eta  + \rho \xi \right)
\end{aligned}
\label{rossler_ou}
\end{equation}
where $\xi(t)$ is Gaussian white noise, and the parameters $\tau_c$ and $\rho$ control the correlation time and amplitude of the OU process, respectively. 
The resulting system has a time-varying, stochastic control parameter, making the learning task harder. 
We use the trajectories generated by this noise-driven system as training data for a single reservoir model.

\begin{figure}[htbp]
    \centering
    \includegraphics[width=0.42\textwidth]{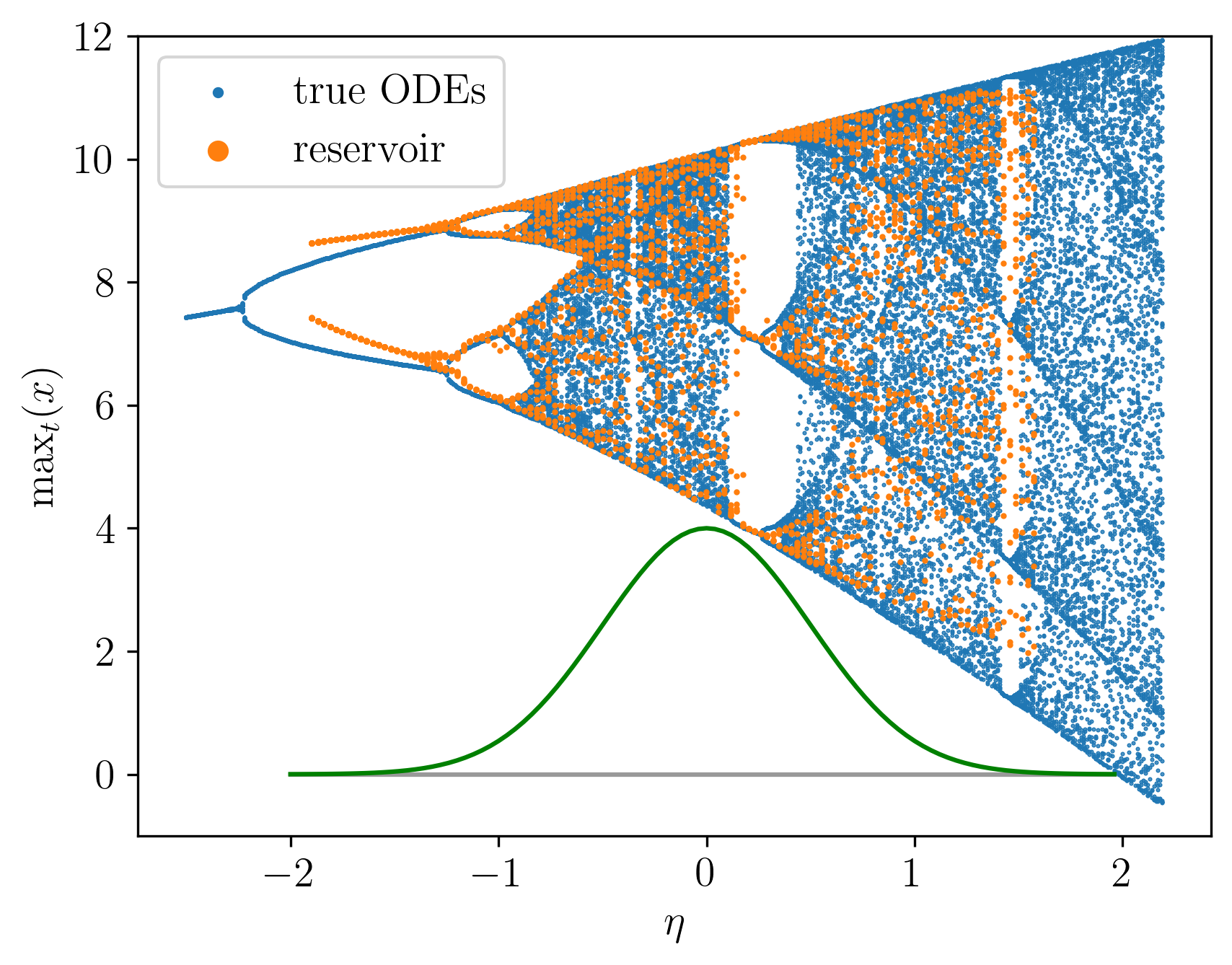}
    \includegraphics[width=0.42\textwidth]{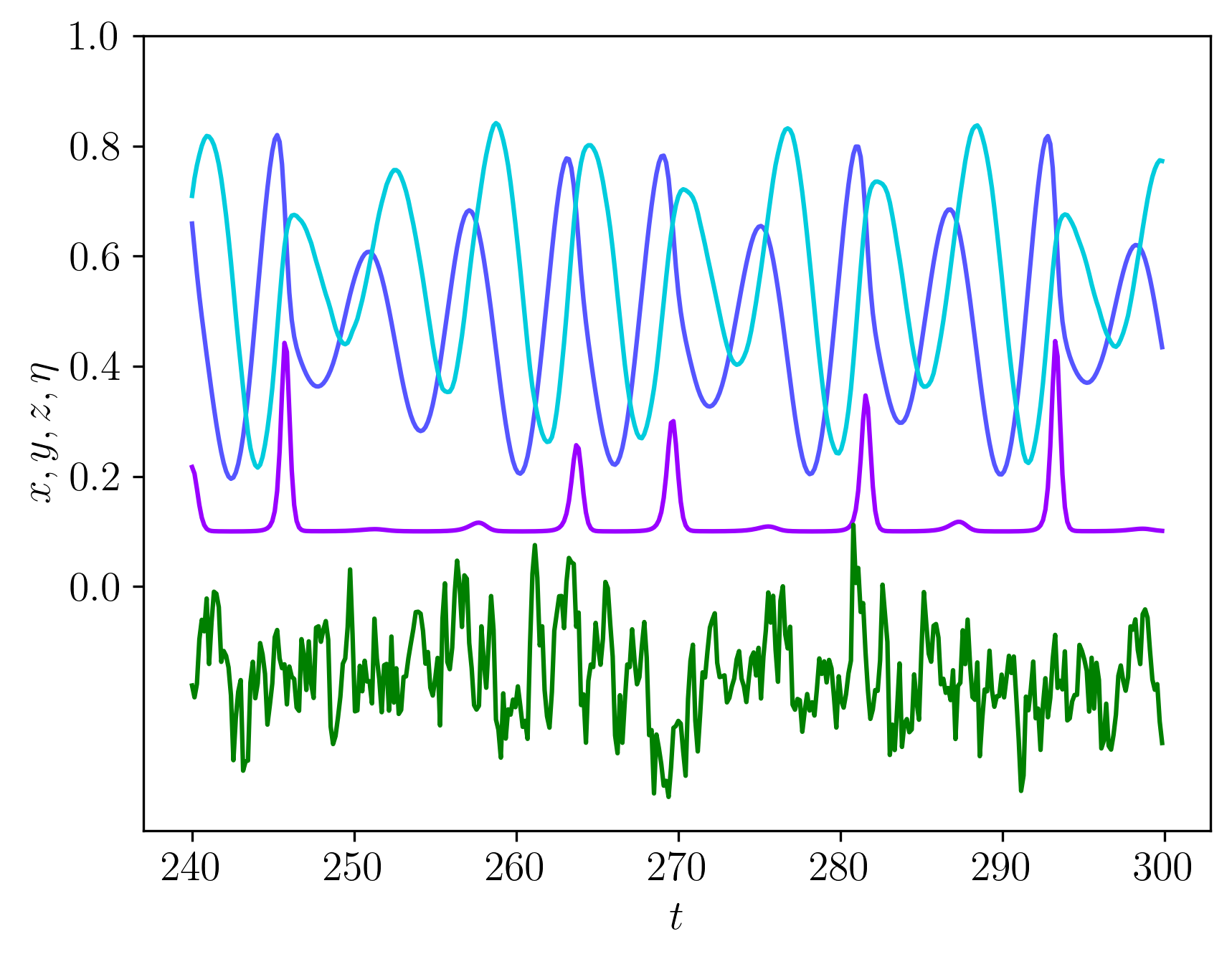}
    \caption{Top panel: bifurcation diagram of the true Rössler system~\eqref{rossler} (blue) and the reconstruction from the reservoir (orange). 
    The training data were obtained from a Rössler system perturbed by an Ornstein-Uhlenbeck input (Eq.~\eqref{rossler_ou}). 
    The green curve shows the empirical probability density of the noise variable $\eta$. 
    The bifurcation points are computed by plotting the local maxima of $x(t)$. 
    Training used a long trajectory ($T = 10^5$), all three variables ($x,y,z$), reservoir size $M = 6000$, delay dimension $H = 25$, and 1\% measurement noise. 
    Bottom panel: the training signal, Rössler variables $(x,y,z)$ are shown in blue, cyan, and purple, and the OU input $\eta(t)$ in green (vertically shifted for clarity). }
    \label{biffurcation2}
\end{figure}

In this case as well, the reservoir accurately reproduces the bifurcation structure near trained values of $\eta$, demonstrating that even from trajectories with rapidly time-varying control parameters, it can learn the correct asymptotic dynamics corresponding to fixed parameter values. Deviations arise primarily when extrapolating beyond the training range, see Fig.~\ref{biffurcation2}. 
This again reflects the fact that the model is asked to infer behavior for parameter values that are only sparsely or indirectly represented in the training data.

\subsection{Estimation of asymptotic phases}\label{sec:ex3}
We again consider the noisy Rössler system~\eqref{rossler_ou}, however, now in a parameter regime where the deterministic system exhibits a stable period-1 limit cycle: $a = 0.2$, $b = 1.6$, $c = 5.7$.
We train a reservoir model on time series of the variables $x(t)$, $y(t)$, and $z(t)$. Unlike in the bifurcation diagram experiment (Sec.~\ref{sec:bifurcation}), we do not provide the noise signal $\eta(t)$ as an explicit input to the reservoir. 
Instead, $\eta(t)$ only perturbs the training trajectory (and is thus hidden from the reservoir input), introducing variability that helps the data explore the state space more broadly. The reservoir is expected to ``average out'' this noise and approximate the dynamics of the unperturbed Rössler system over a wider range of states.

\begin{figure}[htbp]
    \centering
    \includegraphics[width=0.49\textwidth]{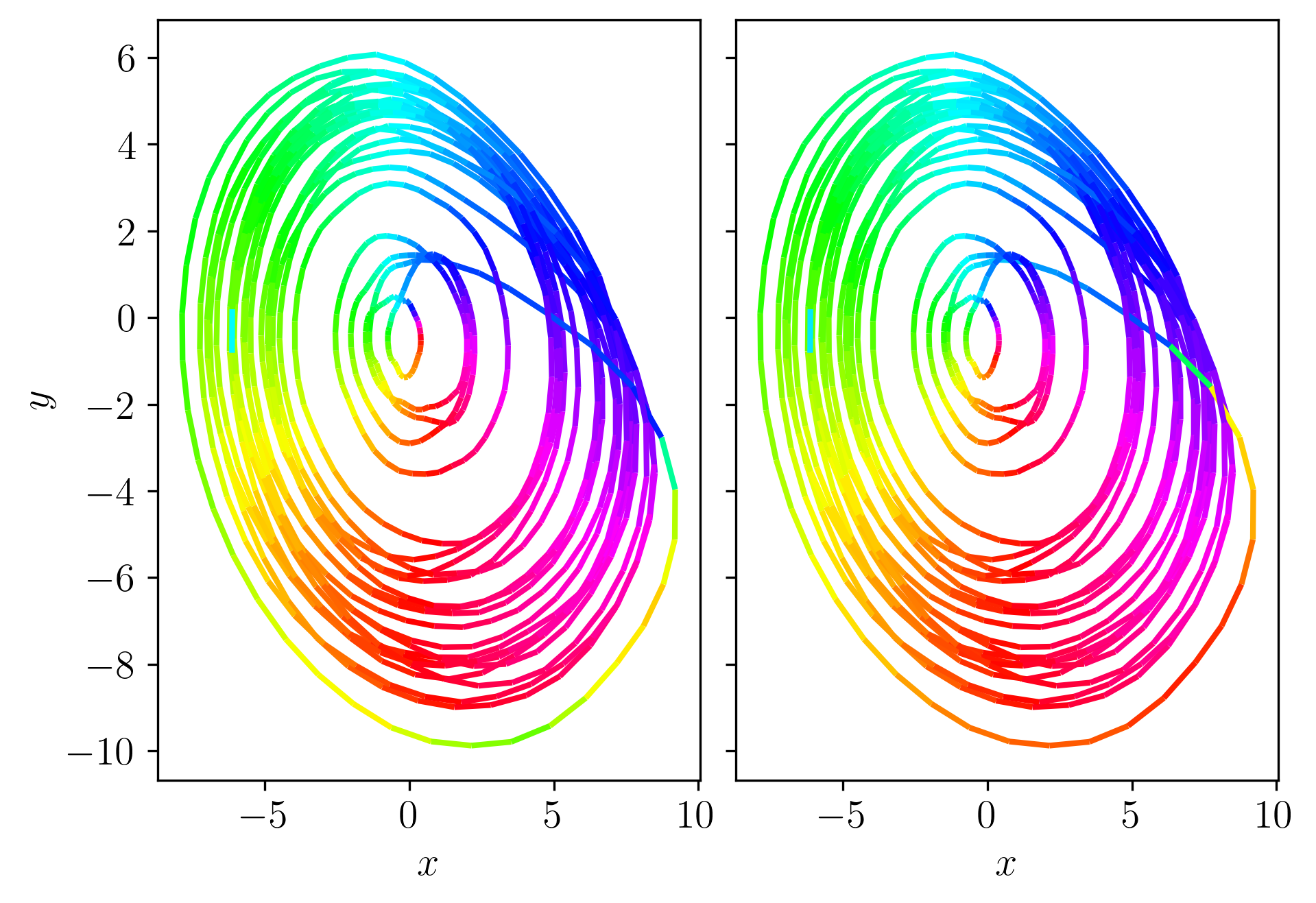}
    \caption{Asymptotic phase estimated along a test trajectory not used during training. Left: true phase obtained by evolving each state forward using the Rössler ODE system~\eqref{rossler} until convergence to the limit cycle. Right: phase estimated by evolving the same states with the trained reservoir model. Color indicates the recovered phase. The reservoir accurately captures the phase structure, except in regions that were poorly sampled during training (see, for example, the rare trajectory excursion in the lower region of the state space). }
    \label{fig:phase}
\end{figure}

We then evaluate the model's ability to recover asymptotic phase. This is a hard task to accomplish, as it requires coherent long-term prediction and a globally accurate model of the evolution of the system.
To test this, we generate a new trajectory from the noisy Rössler system~\eqref{rossler_ou} --- one that has not been used in the training phase --- to use as test data. For each point along this trajectory, we evolve the system forward using the reservoir model until it converges to the limit cycle, and then estimate its asymptotic phase from the position on the cycle. The same is done with the true ODEs for comparison.
We visualize the estimated phases along the test trajectory using a color palette. The reservoir accurately recovers the phase structure, except in regions of state space that were underrepresented in the training data; see Fig.~\ref{fig:phase}.

\section{Discussion}\label{sec:discussion}

In our work, we built on the next-generation reservoir computing framework (NGRC) and proposed a projection method based on pseudorandom nonlinear features. An advantage of this approach is that the number of projected features $M$ can be set independently of the input dimension, whereas in polynomial-based projections the feature space typically depends on the input size $N\cdot H$. We demonstrated the utility of this scalable scheme across several canonical examples, including attractor reconstruction, bifurcation diagram estimation, and asymptotic phase inference. The models showed robust performance even when trained on limited or non-stationary data, and they could generalize to parameters and dynamics not seen during training. These features suggest that next-generation reservoir computing may be a practical tool for applications ranging from digital twins and surrogate modeling to problems in climate science, finance, and biology. 
Reservoir computing has also been applied to real-world forecasting problems, such as anticipating food price fluctuations in agricultural markets~\cite{DOMINGO2023}.

One of the central advantages of NGRC over classical reservoir computing lies in its transparency and manipulability. 
Because NGRC evolves entirely in the space of observable inputs, it allows direct and global perturbations of the system state within this observable space (i.e., by modifying the delay-embedded measurement vector \( \uu_H \)).
This feature is particularly valuable for digital twin and surrogate modeling tasks, where probing system responses to hypothetical inputs or counterfactual states (i.e., hypothetical system states not visited in the observed trajectory) is essential.
In contrast, classical RC relies on evolving a high-dimensional reservoir state \( \rr(t) \), which is difficult to interpret or initialize, making it challenging to explore off-attractor or perform controlled interventions.

Another strength of NGRC is its use of explicit time-delay embedding, which eliminates the need for long transients to ``wash out'' initial conditions, as required in classical RC.
This approach makes it easier to analyze the model, set initial states, and perform  controlled simulations.
Even though the projection \( P(\uu_H) \) may involve a black-box transformation, the evolution remains anchored in physically meaningful variables, enhancing interpretability and stability.

A known challenge in NGRC, highlighted in Ref.~\cite{ratas_pyragas_2024}, is transversal instability in long-term autonomous prediction, whereby small perturbations orthogonal to the attractor can grow and lead to rapid divergence. We examined this issue directly and found that, although Tikhonov regularization helps mitigate overfitting and numerical ill-conditioning in one-step regression
--- consistent with conditioning effects reported in Ref.~\cite{zhang2025chaos} ---
it does not prevent long-term instabilities: models trained without noise generally diverge quickly when iterated autonomously, even when regularized. By contrast, adding a small amount of measurement noise to the training inputs reliably suppresses such failures (similar stabilizing effects were reported in classical RC~\cite{wikner_et_al_ott_2024}). 
Post-processing approaches for enhancing long-horizon robustness have also been considered~\cite{berry_das_2025}.
Across all systems tested, we did not observe any transverse instability once 1\% measurement noise was included, even for extremely long autonomous rollouts used in our quasi-continuation bifurcation reconstructions. A plausible explanation is that noise spreads the training samples slightly off the attractor surface, so the regression must learn a mapping that remains locally consistent not only along the attractor but also in nearby transverse directions (see also the structural stability considerations discussed in Ref.~\cite{santos2025arxiv}). 
Deterministic trajectories sample only a thin manifold and leave transverse directions weakly constrained; noise broadens this sampling, which empirically yields models that do not amplify small off-manifold deviations. We therefore use 1\% input noise as our default configuration in all experiments. 

One promising direction for future work is to apply NGRC to network inference.
By training NGRC models on multivariate time series from interacting systems, one can construct a complete model of the network, which can then be interrogated using surrogate modeling or digital twin techniques to estimate both the intrinsic dynamics of individual units as well as their functional connectivity patterns.

\section*{Data Availability Statement}
Data sharing is not applicable to this article as no new data were created or analyzed in this study. 
For reproducibility, all code used in this work is publicly available at \url{https://github.com/rok-cestnik/next-gen-res-comp}, Ref.~\cite{code}.

\begin{acknowledgments}
We thankfully acknowledge useful discussions with Irmantas Ratas, Andrew Flynn, Edmilson Roque dos Santos and Herbert Jaeger. 
We acknowledge financial support from the Crafoord Foundation (project no. 20240689). 
\end{acknowledgments}

\end{document}